\documentclass{article}

\usepackage{microtype}
\usepackage{graphicx}
\usepackage{subfigure}
\usepackage{booktabs} 

\usepackage{hyperref}

\usepackage[accepted]{icml2024}

\usepackage{pifont}

\usepackage{epsfig}
\usepackage{amsmath}
\usepackage{enumitem}
\usepackage{tabu}
\usepackage{tabularx}
\usepackage{diagbox}
\usepackage{threeparttable}
\usepackage{amsthm}
\usepackage{amssymb}
\usepackage{amsfonts}
\usepackage{mathtools}
\usepackage{multirow}
\usepackage{float}
\usepackage{bm}
\usepackage{array}
\usepackage{xcolor}
\PassOptionsToPackage{table}{xcolor} 
\definecolor{mygray}{gray}{1}
\definecolor{mygray1}{gray}{.90}
\definecolor{reda}{RGB}{202,0,0}
\definecolor{redb}{RGB}{217,148,143}
\definecolor{myyellow}{RGB}{190,144,0}
\definecolor{mygreen}{RGB}{0,136,51}
\definecolor{myblue}{RGB}{0,102,204}
\definecolor{mycomment}{RGB}{142, 193, 229}
\definecolor{myblue}{RGB}{218,232,252}
\definecolor{demphcolor1}{gray}{.6}
\usepackage{colortbl}

\usepackage[capitalize,noabbrev]{cleveref}

\theoremstyle{plain}

\theoremstyle{definition}

\theoremstyle{remark}

\usepackage[textsize=tiny]{todonotes}
\usepackage{amsmath}
\usepackage{amssymb}
\usepackage{mathtools}
\usepackage{amsthm}

\usepackage[capitalize,noabbrev]{cleveref}

\theoremstyle{plain}
\theoremstyle{definition}
\theoremstyle{remark}

\usepackage[textsize=tiny]{todonotes}

\icmltitlerunning{Spider: A Unified Framework for Context-dependent Concept Segmentation}

\begin{document}

\twocolumn[
\icmltitle{Spider  \makebox[14pt][r]{\raisebox{-0.55ex}{\includegraphics[scale=0.2]{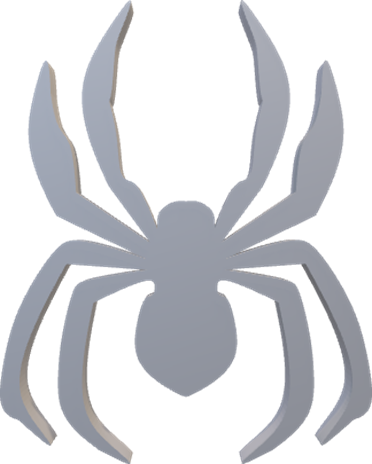}}}: A Unified Framework for Context-dependent Concept Segmentation}



\icmlsetsymbol{equal}{*}

\begin{icmlauthorlist}
	\icmlauthor{Xiaoqi Zhao}{equal,yyy,comp}
	\icmlauthor{Youwei Pang}{equal,yyy,comp}
	\icmlauthor{Wei Ji}{equal,sch}
	\icmlauthor{Baicheng Sheng}{yyy}
	\icmlauthor{Jiaming Zuo}{comp}
	\icmlauthor{Lihe Zhang}{yyy}
	\icmlauthor{Huchuan Lu}{yyy}
\end{icmlauthorlist}

\icmlaffiliation{yyy}{Dalian University of Technology, China}
\icmlaffiliation{comp}{X3000 Inspection Co., Ltd, China}
\icmlaffiliation{sch}{Yale University, America}

\icmlcorrespondingauthor{Lihe Zhang}{zhanglihe@dlut.edu.cn}
\icmlkeywords{Machine Learning, ICML}

\vskip 0.3in
]



\printAffiliationsAndNotice{\icmlEqualContribution} 

\begin{abstract}
	Different from the context-independent (CI) concepts such as human, car, and airplane, context-dependent (CD) concepts require higher visual understanding ability, such as camouflaged object and medical lesion.
	Despite the rapid advance of many CD understanding tasks in respective branches, the isolated evolution leads to their limited cross-domain generalisation and repetitive technique innovation.
	Since there is a strong coupling relationship between foreground and background context in CD tasks, existing methods require to train separate models in their focused domains.
	This restricts their real-world CD concept understanding towards artificial general intelligence (AGI). We propose  a unified model with a single set of parameters, Spider, which only needs to be trained once.
	With the help of the proposed concept filter driven by the image-mask group prompt, Spider is able to understand and distinguish diverse strong context-dependent concepts to accurately capture the Prompter's intention. Without bells and whistles, Spider significantly outperforms the state-of-the-art specialized models in 8 different  context-dependent segmentation tasks, including 4 natural scenes (salient, camouflaged, and transparent objects and shadow) and 4 medical lesions (COVID-19, polyp, breast, and skin lesion with color colonoscopy, CT, ultrasound, and dermoscopy modalities).
	Besides, Spider shows obvious advantages in continuous learning.
	It can easily complete the training of new tasks by fine-tuning parameters less than 1\%
	and bring a tolerable performance degradation of less than 5\% for all old tasks. The source code will be publicly available at \href{https://github.com/Xiaoqi-Zhao-DLUT/Spider-UniCDSeg}{Spider-UniCDSeg}.
\end{abstract}

\begin{figure}[t] 
	\centering
	\includegraphics[width=\linewidth]{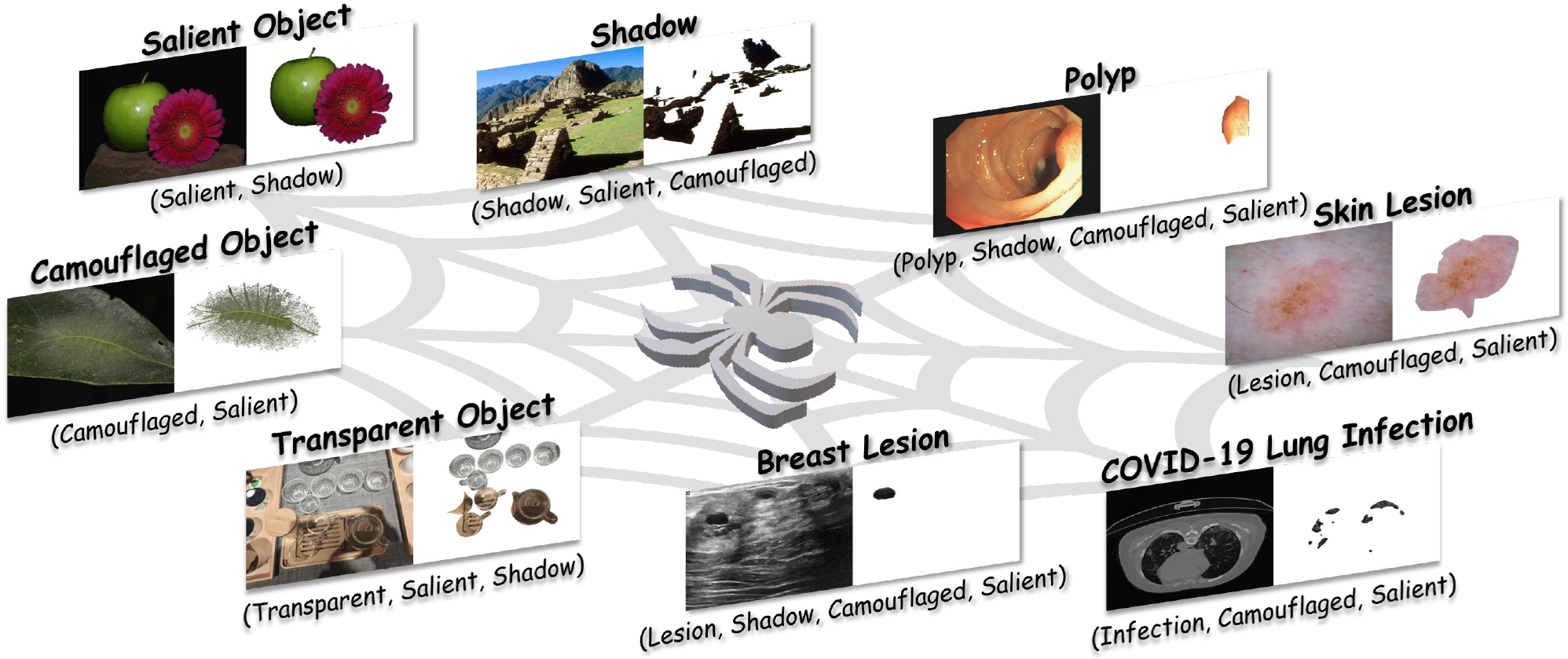}
	\vspace{-5mm}
	\caption{Eight different segmentation tasks with context-dependent concepts are unified into our Spider model. With the interlaced concepts within task domains and class semantic space, Spider can wander to any target of interest.
	}
	\label{fig:task_spider}
	\vspace{-5mm}
\end{figure}

\section{Introduction}
In philosophy and cognitive science~\cite{CICD_1,CICD_2,CICD_3}, concepts usually contain the context-independent (CI) and  context-dependent (CD) concepts. 
For example, semantic segmentation tasks define numerous CI concepts (i.e., multi-granularity semantic classes). The class of an object is fixed no matter what scene it locates in.
The context-dependent concepts mean that the target is not cognizable without its environment,
such as salient/camouflaged object detection, shadow detection, medical lesion segmentation, as shown in~\cref{fig:task_spider}.
People determine where the target locates mainly according to the surroundings.
In this work, we focus on the context-dependent segmentation tasks and expect to build a parameter-unified framework.
Existing works explore the in-domain modeling, resulting in repetitive structure design, inefficient data utilisation, and limited multi-domain generalisation.
As an alternative, a seminal work of EVP~\cite{EVP} attempts to unify three CD tasks based on low-level structure, but it still requires to train the model one by one for each task and lacks the unification of parameters.

With the development of strong backbones like ConvNeXt~\cite{ConvNeXt} and ViT\cite{ViT}, and visual prompt technology~\cite{UNINEXT,PromptIR,Painter,SegGPT}, some unification models appear towards the attempt of AGI.
In the field of segmentation, generalist models typified by SAM~\cite{SAM} and SegGPT~\cite{SegGPT}, have shown gratifying versatility.
They rely on the visual prompt of a single pair of image and foreground  to understand CI concepts.
However, many reports~\cite{SAM_report1,SAM_report2,SAM_report3,SAM_report4,SAM_report5,SegGPT_report1,ji2023segment} show their poor performance on CD concept understanding in terms of both quantitative and qualitative evaluations.
It is because the targets in the CD tasks have not the fixed semantic classes, and multiple CD concepts often mingle together in semantic space (Please see~\cref{fig:task_spider}, in which one target exhibits multiple CD properties and one CD concept covers multiple semantic classes),
thus the prompt of a single foreground  fails to provide explicit guidance for the segmentation model. 
To design a parameter-unified context-dependent segmentation model, the first critical issue is how to build a compatible pipeline to understand and distinguish each CD concept.
The second one is how to overcome the challenges posed to cross-domain collaborative learning because of the overlapped semantic classes across the foregrounds of different tasks and the large gap among different task domains.

In this paper, we propose a unified CD concept segmentation framework \textbf{Spider}, which shares 100\% parameters for all tasks based on the idea of dynamic filtering.
For structure unification, Spider equips with a segmentation stream and a concept prompt stream.
When facing each task, the prompt stream generates a concept filter to act on the tail of the segmentation stream, thereby yielding a unified single-channel output.
For parameter unification, the segmentation stream is responsible for learning task-generic representations from cross-domain data through a single set of encoder-decoder parameters.  This stream can be substituted to any specialized models.

\begin{figure}[t]
	\centering
	\includegraphics[width=\linewidth]{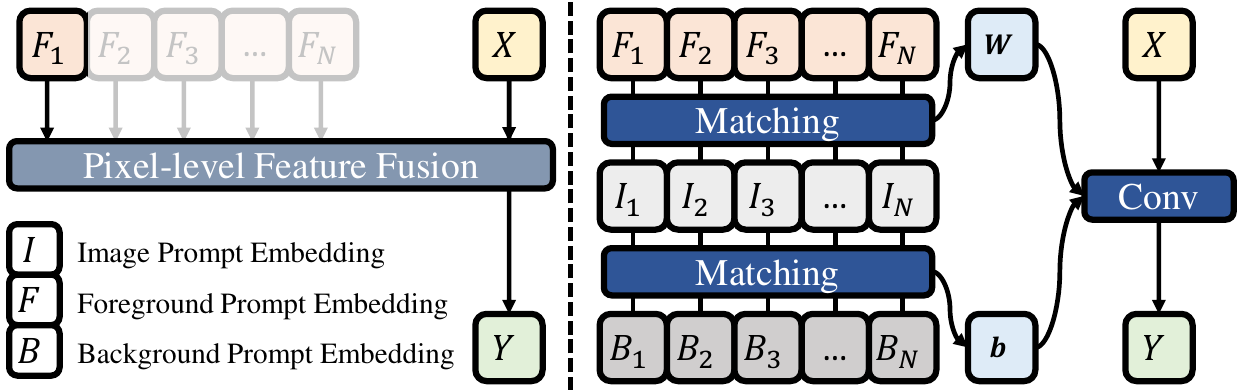}
	\vspace{-5mm}
	\caption{Two types of feature interaction between visual prompts and current image input. The left is used in Universeg~\cite{UniverSeg} with multiple foreground prompts and others~\cite{SegGPT,SAM,clip_seg2} with a single foreground prompt. The right is ours.
	}
	\label{fig:Interaction}
	\vspace{-5mm}
\end{figure}

Previous unified segmentation models~\cite{SAM,SegGPT,UniverSeg,clip_seg2} adopt non-local/co-attention or element-wise spatial operations to propagate foreground prompt knowledge to current input features at pixel  level, as shown in \cref{fig:Interaction}.
Because context-dependent targets have uncertain categories,
pixel-level feature fusion with foreground prompt embeddings easily produces ambiguities in prompt definitions and is more sensitive to the accuracy of mask annotation. Different from them, we utilize a group of image, foreground and background prompts to comprehensively mine the clues of CD  targets and establish high-level image-foreground matching and image-background matching to achieve feature interaction between visual prompt and current input by the concept filter.

Specifically, 
we utilize the transformer~\cite{Transformer} to establish long-range dependence of concepts and environments within a group of images. Image-group prompts are encoded as the key and value. Both foreground and background mask-group prompts are embedded and encoded as the query. Through multiple cross-attention operations, the model can learn the common conceptual expressions in the group of images indicated by their masks. The updated object embeddings and context-aware feature embeddings are used as the weight and bias of concept filter, respectively.
With the help of the concept filters, Spider can wander across different task domains and establish the connections among these CD concepts. It supports customized prompts during inference and has the potential of perceiving unseen context-dependent objects.
More advantages of the concept filter have been summarized in the \cref{sec:advantages_CF}.
In addition, we devise a ``Balance FP - Unify BP'' training strategy to balance different tasks in both forward-propagating batch specification and back-propagating gradient update, thereby guaranteeing the performance in all tasks.

Our main contributions can be summarized as follows:
\begin{itemize}[noitemsep,nolistsep]
	\item We propose a unified model, Spider, which only needs to be trained once and can perform complex context-dependent concept understanding in diverse domains.
	\item Benefiting from the flexible concept filter, Spider can sensitively perceive the attributes or categories of interest, so that it can be trained on different domains without heavy task-specific heads.
	\item Spider achieves superior performance in 8 challenging tasks with context-dependent concepts,
	and it  has powerful continuous learning abilities.
	It can be generalized to new tasks by fine-tuning parameters of less than 1\% without any structural modifications, and preserve the performance on old tasks with slight  degradation of less than 5\%. As the scale and diversity of training data increase, it shows the potential in unseen tasks.
\end{itemize}

\section{Related Work}
\begin{figure}[t]
	\centering
	\includegraphics[width=\linewidth]{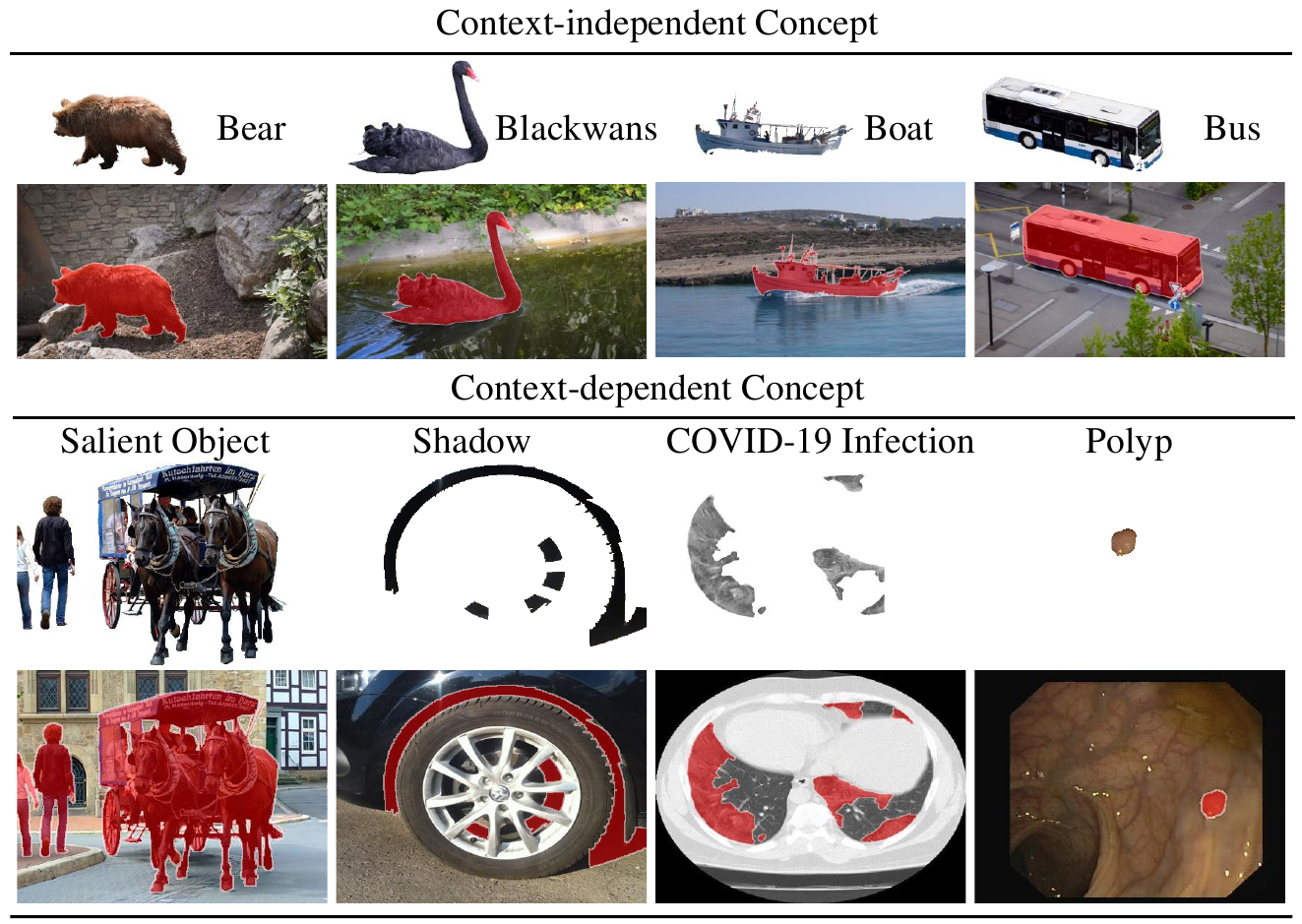}
	\vspace{-5mm}
	\caption{Visual comparison of segmentation objects with context-independent concepts and  context-dependent concepts. The odd rows are the pure foregrounds and the even rows are the complete images with the \textcolor{red}{highlighted} foregrounds.}
	\label{fig:context_de_vs_inde}
	\vspace{-5mm}
\end{figure}
\subsection{Context-dependent Image Segmentation}

As shown in~\cref{fig:context_de_vs_inde}, the context-independent concept can be well understood without the help of its contexts.
While the context-dependent concept is the complete opposite, we have to rely on its surroundings for a clear understanding.
In this work, we choose eight representative tasks, including four natural scene tasks and four different modality medical tasks, to investigate the unified modeling for the context-dependent concept segmentation.
Detailed definition of these context-dependent segmentation tasks can be found in the \cref{sec:def_CIS}.
Among these context-dependent segmentation tasks, the U-shaped structure~\cite{UNet,FPN} with the encoder-decoder form is the most basic framework.
According to the characteristics of different tasks, existing methods mainly focus on four aspects:
{visual attention}~\cite{GateNet,BDRAR_Shadow,ACSNet_Polyp,UACANet_Polyp,li2023delving,li2021joint,ji2023multispectral},
{multi-scale feature extraction}~\cite{MINet,ZoomNet_COD,DilatedConvolution-TKCN,DilatedConvolution-D3DNet,DANet_RGBDSOD,li2023dvsod,ji2022promoting,piao2019depth},
{edge refinement}~\cite{BGNet_COD,SegMaR_COD,PMD_Mirror,CFANet_Polyp,RCARP_Glass},
and {combination of different architectures}~\cite{TransFuse_Polyp,Trans2Seg_Transparent,UGTR_COD,VST,ji2021learning}.
In this work, we consider the cross-domain learning and design an efficient and unified framework with only one set of parameters and one training session.

\begin{figure*}[!t]
	\centering
	\includegraphics[width=\linewidth]{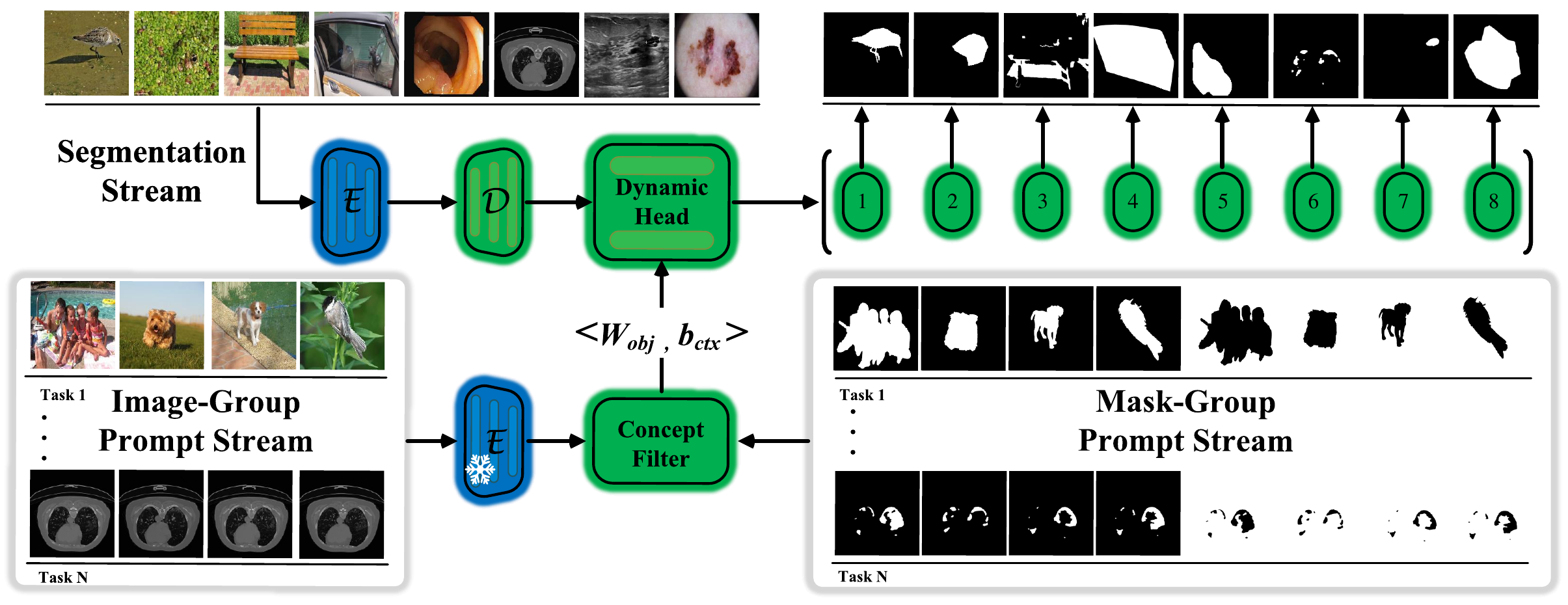}
	\vspace{-5mm}
	\caption{Overall pipeline. It consists of  segmentation stream $\mathcal{S}_{s}$, and image- and mask- group prompt streams $\mathcal{S}_{i}$ and $\mathcal{S}_{m}$.
		$\mathcal{S}_{s}$ uses the encoder-decoder structure.
		$\mathcal{S}_{i}$ is fed into the frozen pre-trained encoder and output the group prompt feature $F_{mem}$ as the key and value of the transformer decoder.
		$\mathcal{S}_{m}$ generates the foreground-aware and background-aware queries by masked average pooling on the group prompt features $F_{mem}$.
		A series of concept filters $<W_{obj}, b_{ctx}>$ act on the last layer of the decoder to generate dynamic prediction.
	}
	\label{fig:framework}
	\vspace{-5mm}
\end{figure*}

\subsection{Unified Vision Models}

With the development of foundation models~\cite{Swin,ViT,ConvNeXt,Internimage, Dino}, solving multiple vision tasks by a single set of model parameters has become an important way to move towards AGI. 
Previous typical parameter-unified methods~\cite{MaskR-cnn,Mask2Former,MTI-Net,MMFT} are mainly based on multi-task learning. They design multiple task-specific heads/decoders for different tasks, such as object detection and panoptic/instance/semantic segmentation as in~\cite{MaskR-cnn,Mask2Former}, and SOD and depth/edge estimation as in~\cite{MTI-Net,MMFT}. 
Because each sample has multiple annotations corresponding to different tasks at the same time, all these studies are  performed in-domain and handle different tasks by different heads. 
However, in the real world, different tasks focus on different objects of interest, data domains, and annotation types.
Therefore, the cross-domain learning is a key paradigm to  unify model parameters.
A simple query-based task formulation is proposed in~\cite{UniHCP} for handling multiple distinctly defined human-centric tasks.
Ten instance perception tasks are unified into a prompt-guided object discovery and retrieval paradigm in~\cite{UNINEXT}.
Input-conditioned prompts with the contextual information in~\cite{PromptIR} is designed to guide different image restoration tasks.
In~\cite{Painter}, the task-specific input-output image pair is used as condition to perform ten different dense prediction tasks.
In terms of image segmentation, CLIP-driven universal models~\cite{clip_seg1,clip_seg2} incorporate text embedding to provide the models with different semantic prompts. UniverSeg~\cite{UniverSeg} employs feature fusion with the query image and example set of image-label pairs to achieve universe medical image segmentation. SAM~\cite{SAM} designs a powerful segmentation architecture equipped with the reusable image embedding and a orientated prompt branch. In SegGPT~\cite{SegGPT}, the image segmentation is formulated as an in-context coloring problem, which requires a image-mask pair prompt to indicate object segmentation. However, the motivation of these unified methods is oriented to CI concepts. They only focus on the single pair of image-foreground prompt or embedding the prompt knowledge by the pix-level feature fusion. Many works~\cite{SAM_report1,SegGPT_report1,SAM_report4,SAM_report5,SAM_report6} report that the current unified/generalist methods are still difficult to handle diverse strong context-dependent segmentation tasks during training and inference. In this work, we propose a simple unified architecture guided by image-mask (foreground and background) group prompts for eight CD concept tasks with multiple modalities.

\section{Approach}

To furthest share knowledge among various tasks and reduce specialized designs, we attempt to maximize weight sharing.
As shown in~\cref{fig:framework}, we directly utilize the general encoder-decoder architecture without any modifications, i.e., a vanilla UNet~\cite{UNet} or FPN~\cite{FPN} with different backbones.
Our core component is the concept filters generated by image and mask prompt streams $\mathcal{S}_{i}$ and $\mathcal{S}_{m}$, which are embedded in the final dynamic head to accurately predict different tasks.
In this way, all feature extractions and fusion weights absorb multi-domain information and share 100\% parameters across all tasks.
\subsection{Prompt Generation}
\label{sec:Prompt Generation}

The prompt generation strategy is different in the training and inference period.
During training, firstly, $G$ pairs of images and masks are randomly selected from each task at each iteration as group prompts.
This manner of random combination ensures the performance stability of the concept filter when facing different group prompts in practical applications, and its motivation and effects are similar to those of the masked image modeling (MIM) mechanism~\cite{Beit,MAE,Simmim,SegGPT}.
Next, we pass the image group to the frozen pre-trained encoder $\mathbf{E}$ to obtain rich high-level semantic features.
Finally, the concept filters $<W_{obj}, b_{ctx}>$ derived from the image-group features $F_g$ and mask-group maps $M_{fg}$ and $M_{bg}$, participate in the convolutional dynamic head for prediction.
During inference, to ensure stability and replicability of predictions, we look through all training samples and select $64$ representative examples for each task as its fixed group prompt by K-means clustering over their high-level embeddings. Specifically, we first fed all images to the encoder of prompt stream. The extracted high-level feature maps are global average pooled to condense semantic information and reduce the computational complexity in the clustering process.
Next, we randomly generate $64$ initial clustering centers, iteratively cluster high-level embeddings based on the similarity, and update these centers until convergence. Finally, we select the samples closest to cluster center as the image-group prompts.
Quantitative results can be seen in~\cref{tab:ablation_study} and the visualization of clustered group prompt for each concept is presented in the~\cref{sec:vis_cgp}.
It is worth noting that group prompts can be flexibly provided by users and are not limited to these clustered prompts in practical applications.

\begin{figure*}[t]
	\centering
	\includegraphics[width=\linewidth]{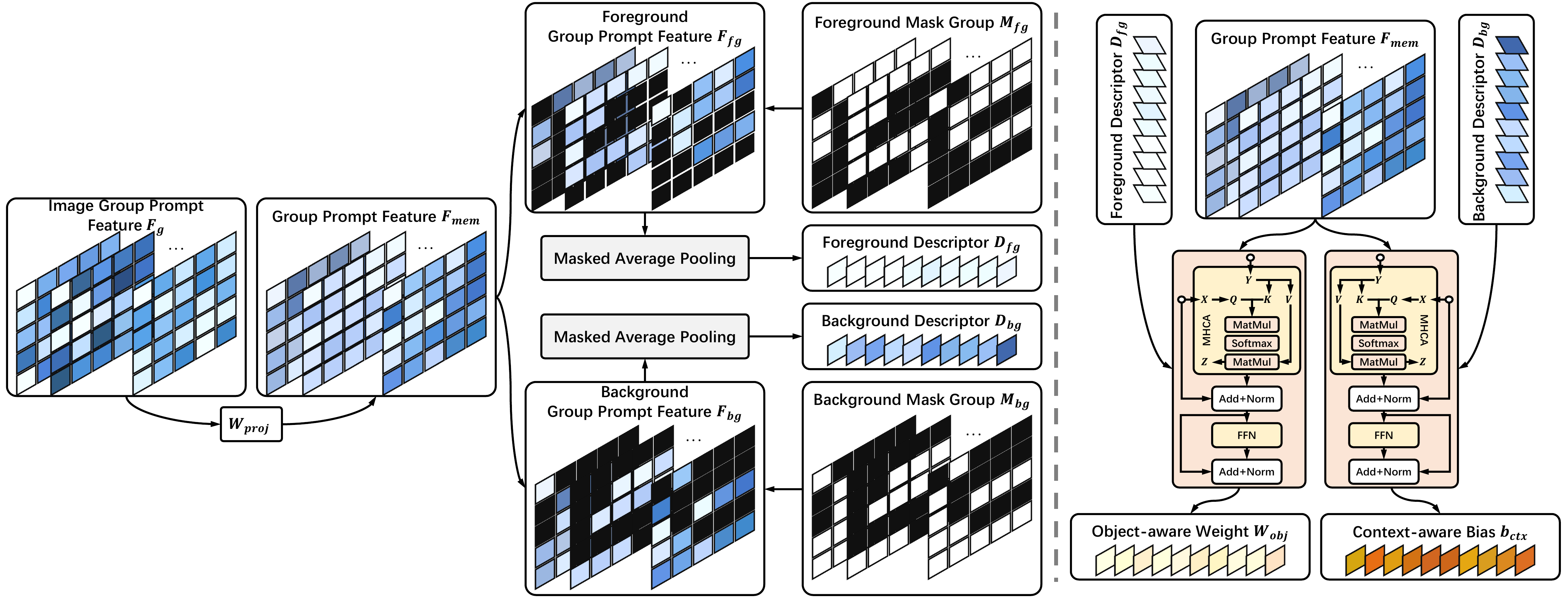}
	\vspace{-5mm}
	\caption{Illustration of generating concept filters.}
	\label{fig:filter}
\end{figure*}

\subsection{Concept Filter}
This component is the key. It unifies multiple tasks into a single framework through the idea of conditional filtering. The details are shown in~\cref{fig:filter}.
Specifically, we use the learnable projection matrix $W_{proj}$ to transform the deep representations $F_g$ of the image-group prompt and obtain the group prompt feature $F_{mem}$.
The foreground mask group $M_{fg}$ and background mask group $M_{bg}$ corresponding to the targets of interest in the image-group prompt guide to yield foreground descriptor $D_{fg}$ and background descriptor $D_{bg}$ by  masked average pooling.
This extracts rough representations about foreground and background from $F_{mem}$  specific to the contexts of current task.
We further refine the two descriptors by mining foreground/background related semantic cues in the global context from appearance-driven $F_{mem}$.
This process is achieved through multi-head cross-attention (\texttt{MHCA}).
In \texttt{MHCA}, $D_{fg}$ and $D_{bg}$ separately act as $X$, which is further linearly transformed to $Q$.
And $F_{mem}$ serving as $Y$ are mapped to $K$ and $V$ as well.
The foreground or background activation map $M$ is computed as:
\begin{equation}
\begin{split}
M = \texttt{softmax}(\frac{Q K^{\top}}{d}),
\end{split}
\label{equ:mhca}
\end{equation}
where $d$ is a normalization factor.
$M$ is exploited to aggregate contextual information from $V$.
\begin{equation}
\begin{split}
Z = X + M V W_Z, \\
X = Z + \texttt{FFN}(Z), \\
\end{split}
\label{equ:ffn}
\end{equation}
where  $W_Z$ is the learnable weight. %
After the cascaded \texttt{FFN}, the resulting foreground and background descriptors are taken as object-aware weight $W_{obj}$ and context-aware bias $b_{ctx}$ of concept filter, respectively.

\begin{algorithm}[!t]
	\small
	\caption{Training and Inference}
	\label{alg:un}

	\textit{Training iteration with $N=16$ and $B=4$.}
	
	\begin{algorithmic}[1]
		\REQUIRE A batch $D = \{D_{t}\}_{t=1}^{8}$. $D_{t}$ is $N$ image-mask pairs randomly selected from training set of task $t$.

		\STATE create image tensor $I \in \mathbb{R}^{8 \times N \times 3 \times H \times W}$ from $D$   
		\STATE create mask tensor $M \in \mathbb{R}^{8 \times N \times 1 \times H \times W}$ from $D$
		\FOR{$i \gets 0, N/B - 1$}
		\STATE $f \gets \mathbf{D}(\mathbf{E}(I[:, iB:(i+1)B, ...] ))$
		\hfill \textcolor{mycomment}{$//$ generate image feature}
		
		\STATE $L \gets 0$
		\FOR{$t \gets 1, 8$}
		
		\STATE $P_t^I \gets Cat(I[t, 0:iB, ...],I[t, (i+1)B : N, ...])$  
		\hfill \textcolor{mycomment}{$//$ image group prompt}
		
		\STATE $P_t^M \gets Cat(M[t, 0:iB, ...],M[t, (i+1)B : N, ...])$
		\hfill \textcolor{mycomment}{$//$ mask group prompt}
		\STATE $<W_{obj}, b_{ctx}> \gets \texttt{PromptStream}(P_t^I, P_t^M)$

		\STATE $P_{t} \gets \texttt{DHead}(f[t, ...], <W_{obj}, b_{ctx}>)$

		\STATE $L \gets L + \texttt{Loss}(P_{t}, M[t, iB:(i+1)B, ...])$   
		\ENDFOR
		\STATE \texttt{backward}($L$)   
		\ENDFOR
		\hfill \textcolor{mycomment}{$//$ update parameters}
		
	\end{algorithmic}
	\hrulefill
	
	\textit{Inference with the minibatch of $N$.}
	
	\begin{algorithmic}[1]
		\REQUIRE concept filter set $\{<W_{obj}, b_{ctx}>_{t}\}_{t=1}^{8}$, 
		input tensor $I \in \mathbb{R}^{N \times 3 \times H \times W}$,
		task indicator $t \in \{1,2,...,8\}^N$.
		\ENSURE prediction  $Y \in \mathbb{R}^{N \times 1 \times H \times W}$
		
		\STATE $f \gets \mathbf{D}(\mathbf{E}(I))$

		\FOR{$n \gets 1, N$}
		\STATE $Y[n, \dots] \gets \texttt{DHead}(f, <W_{obj}, b_{ctx}>_{t(n)})$
		\ENDFOR
		\hfill \textcolor{mycomment}{$//$ traverse all images}
	\end{algorithmic}
\end{algorithm}

\subsection{Training and Inference}
To simultaneously balance the performance of these tasks in both forward propagation and back propagation during training, we design a \textbf{``Balance FP - Unify BP''} strategy.
Specifically, we first randomly select $N$ samples for each task, 
of which $B$ samples are input to the segmentation branch and the rest $N - B$ samples are used as prompts.
The samples of all tasks will be concatenated together to separately obtain the input tensor of segmentation stream and prompt stream with a shape of $[8B, C, H, W]$ and $[8(N-B), C, H, W]$. 
During the forward propagation, the batch normalization~\cite{BatchNorm} can make the input distribution of each task closer, which helps the model learn task-shared representations, improving and \textbf{balancing} the overall performance. 
Moreover, we circularly generate eight concept filters in the tail to complete the predictions for the corresponding tasks, avoiding repeated computation caused by full forward propagation. 
Next, we use the PPA loss~\cite{F3Net,PraNet_Polyp,MENet,FEDER,PGNet,Inf-Net,EVP} widely adopted in segmentation tasks to jointly calculate the loss of all samples. 
During the back propagation, the direction of parameter optimization is \textbf{unified} to help Spider obtain better overall performance without favoring a single task. 
In the inference phase, the input of the segmentation stream supports splicing multiple samples in the batch dimension. We may flexibly assign the concept filters to them for single or multiple concept predictions. 
Each concept filter receives a group of customized or fixed prompts from the training set as mentioned in~\cref{sec:Prompt Generation}. The detailed training and inference process can be found in \cref{alg:un}.

\begin{table}[!t]
	\centering
	\caption{
		Data partition in eight tasks, which is widely used by the state-of-the-art specialized methods.}
	\resizebox{\linewidth}{!}{
		 \begin{tabular}{l|l|c|c}
\toprule[2pt]
 Task&Dataset&\#Train&\#Test\\
 \hline
 Salient Object Detection (SOD)&DUTS~\cite{DUTS}&10,548&5,017\\
 Camouflaged Object Detection (COD)&COD10K~\cite{SINet_COD}&4,040 &2,026\\
 Shadow Detection (SD)&SBU~\cite{SBU} &4,085&638\\
 Transparent Object Segmentation (TOS) &Trans10K~\cite{Trans10K}&5,000 &4,428\\	
Colon Polyp Segmentation (CPS) & Five datasets~\cite{PraNet_Polyp} &1450 &798\\
COVID-19 Lung Infection (CLI)&COVID-19 data~\cite{Inf-Net}&894&383\\
Breast Lesion Segmentation (BLS)&BUSI~\cite{BUSI}&486&161\\
Skin Lesion Segmentation (SLS) &ISIC18~\cite{ISIC18}&1,886 &808\\	
\bottomrule[2pt]
\end{tabular}

	}
	\label{tab:Datasets_survey}
	\vspace{-5mm}
\end{table}

\begin{table*}[!t]
	\centering
	\caption{Quantitative comparisons with the unified models 
		and state-of-the-art specialized models on the eight tasks.
		$\uparrow$ and $\downarrow$ indicate that the larger scores and the smaller ones are better, respectively.
		The best scores are highlighted in {\color{reda} \textbf{red}}. Following SegGPT~\cite{SegGPT}, we first adopt ViT-B/L as the encoder of Spider. To facilitate future research comparisons, we further provide the Swin-B/L and ConvNeXt-B/L versions. Our largest version, Spider-ConvNext-L, has the same 1.5G model size as SegGPT.
	}
	\resizebox{\linewidth}{!}{
		\setlength\tabcolsep{2pt}
		\renewcommand\arraystretch{1.2}
		 \begin{tabular}{r|c|c|cc|cc|cc|cc|cc|cc|cc|cc}
\toprule[2pt]

   &   &  
   & \multicolumn{2}{c|}{\large\textbf{{Salient}}}
   & \multicolumn{2}{c|}{\large\textbf{{Camouflaged
}}}
   & \multicolumn{2}{c|}{\large\textbf{{Shadow}}}
   & \multicolumn{2}{c|}{\large\textbf{{Transparent}}}
   & \multicolumn{2}{c|}{\large\textbf{{Polyp}}}
      & \multicolumn{2}{c|}{\large\textbf{{COVID-19}}}
    & \multicolumn{2}{c|}{\large\textbf{{Breast}}}
      & \multicolumn{2}{c}{\large\textbf{{Skin}}}
      \\
   &   &  
   & \multicolumn{2}{c|}{{{DUTS}}}
   & \multicolumn{2}{c|}{{{COD10K
}}}
   & \multicolumn{2}{c|}{{{SBU}}}& \multicolumn{2}{c|}{{{Trans10K}}}
   & \multicolumn{2}{c|}{{{5 datasets}}}
      & \multicolumn{2}{c|}{{{COVID-19 CT}}}
    & \multicolumn{2}{c|}{{{BUSI}}}
      & \multicolumn{2}{c}{{{ISIC2018}}}
\\
     Method&Publication&Backbone  
     &F$_{\beta}^{\omega} \uparrow$ & S$_m \uparrow$
   &F$_{\beta}^{\omega} \uparrow$ & S$_m \uparrow$  &BER $\downarrow$ &MAE $\downarrow$  &BER $\downarrow$ &MAE $\downarrow$ &mDice $\uparrow$ &mIoU $\uparrow$
 &mDice $\uparrow$ &mIoU $\uparrow$
 &mDice $\uparrow$ &mIoU $\uparrow$
 &mDice $\uparrow$ &mIoU $\uparrow$
     \\
    \hline
    \multicolumn{19}{c}{{\large\textbf{Specialized Models
}}}
    \\
    \hline
MENet~\cite{MENet}
& CVPR’23
&ResNet-50~\cite{ResNet}
&0.8698
&0.9028
&\cellcolor{mygray}{-}
&\cellcolor{mygray}{-}
&\cellcolor{mygray}{-}
&\cellcolor{mygray}{-}
&\cellcolor{mygray}{-}
&\cellcolor{mygray}{-}
&\cellcolor{mygray}{-}
&\cellcolor{mygray}{-}
&\cellcolor{mygray}{-}
&\cellcolor{mygray}{-}
&\cellcolor{mygray}{-}
&\cellcolor{mygray}{-}
&\cellcolor{mygray}{-}
&\cellcolor{mygray}{-}
 \\
 PGNet~\cite{PGNet}
& CVPR’22
&Swin-B~\cite{Swin}
&0.8736
&0.9091
&\cellcolor{mygray}{-}
&\cellcolor{mygray}{-}
&\cellcolor{mygray}{-}
&\cellcolor{mygray}{-}
&\cellcolor{mygray}{-}
&\cellcolor{mygray}{-}
&\cellcolor{mygray}{-}
&\cellcolor{mygray}{-}
&\cellcolor{mygray}{-}
&\cellcolor{mygray}{-}
&\cellcolor{mygray}{-}
&\cellcolor{mygray}{-}
&\cellcolor{mygray}{-}
&\cellcolor{mygray}{-}
 \\
FEDER~\cite{FEDER}
&CVPR’23
&ResNet-50~\cite{ResNet}
&\cellcolor{mygray}{-}
&\cellcolor{mygray}{-}
&0.7155
&0.8196
&\cellcolor{mygray}{-}
&\cellcolor{mygray}{-}
&\cellcolor{mygray}{-}
&\cellcolor{mygray}{-}
&\cellcolor{mygray}{-}
&\cellcolor{mygray}{-}
&\cellcolor{mygray}{-}
&\cellcolor{mygray}{-}
&\cellcolor{mygray}{-}
&\cellcolor{mygray}{-}
&\cellcolor{mygray}{-}
&\cellcolor{mygray}{-}
 \\
FSPNet~\cite{FSPNet}
&CVPR’23
&ViT-B16~\cite{ViT}
&\cellcolor{mygray}{-}
&\cellcolor{mygray}{-}
&0.7347
&0.8470
&\cellcolor{mygray}{-}
&\cellcolor{mygray}{-}
&\cellcolor{mygray}{-}
&\cellcolor{mygray}{-}
&\cellcolor{mygray}{-}
&\cellcolor{mygray}{-}
&\cellcolor{mygray}{-}
&\cellcolor{mygray}{-}
&\cellcolor{mygray}{-}
&\cellcolor{mygray}{-}
&\cellcolor{mygray}{-}
&\cellcolor{mygray}{-}
\\
SILT~\cite{SILT}
&ICCV’23
&PVTv2-B5~\cite{PVTv2}
&\cellcolor{mygray}{-}
&\cellcolor{mygray}{-}
&\cellcolor{mygray}{-}
&\cellcolor{mygray}{-}
&0.0402
&0.0493
&\cellcolor{mygray}{-}
&\cellcolor{mygray}{-}
&\cellcolor{mygray}{-}
&\cellcolor{mygray}{-}
&\cellcolor{mygray}{-}
&\cellcolor{mygray}{-}
&\cellcolor{mygray}{-}
&\cellcolor{mygray}{-}
&\cellcolor{mygray}{-}
&\cellcolor{mygray}{-}
\\
SARA~\cite{SARA}
&CVPR’23
&ConvNeXt-L~\cite{ConvNeXt}
&\cellcolor{mygray}{-}
&\cellcolor{mygray}{-}
&\cellcolor{mygray}{-}
&\cellcolor{mygray}{-}
&0.0429
&0.0333
&\cellcolor{mygray}{-}
&\cellcolor{mygray}{-}
&\cellcolor{mygray}{-}
&\cellcolor{mygray}{-}
&\cellcolor{mygray}{-}
&\cellcolor{mygray}{-}
&\cellcolor{mygray}{-}
&\cellcolor{mygray}{-}
&\cellcolor{mygray}{-}
&\cellcolor{mygray}{-}
\\
EBLNet~\cite{EBLNet}
&ICCV’21
&ResNet-50~\cite{ResNet}
&\cellcolor{mygray}{-}
&\cellcolor{mygray}{-}
&\cellcolor{mygray}{-}
&\cellcolor{mygray}{-}
&\cellcolor{mygray}{-}
&\cellcolor{mygray}{-}
&0.1383
&0.0959
&\cellcolor{mygray}{-}
&\cellcolor{mygray}{-}
&\cellcolor{mygray}{-}
&\cellcolor{mygray}{-}
&\cellcolor{mygray}{-}
&\cellcolor{mygray}{-}
&\cellcolor{mygray}{-}
&\cellcolor{mygray}{-}
\\
RFENet~\cite{RFENet}
&IJCAI’23
&ResNet-50~\cite{ResNet}
&\cellcolor{mygray}{-}
&\cellcolor{mygray}{-}
&\cellcolor{mygray}{-}
&\cellcolor{mygray}{-}
&\cellcolor{mygray}{-}
&\cellcolor{mygray}{-}
&0.1036
&0.0767
&\cellcolor{mygray}{-}
&\cellcolor{mygray}{-}
&\cellcolor{mygray}{-}
&\cellcolor{mygray}{-}
&\cellcolor{mygray}{-}
&\cellcolor{mygray}{-}
&\cellcolor{mygray}{-}
&\cellcolor{mygray}{-}
\\
LDNet~\cite{LDNet_Polyp}
&MICCAI’22
&Res2Net-50~\cite{Res2Net}
&\cellcolor{mygray}{-}
&\cellcolor{mygray}{-}
&\cellcolor{mygray}{-}
&\cellcolor{mygray}{-}
&\cellcolor{mygray}{-}
&\cellcolor{mygray}{-}
&\cellcolor{mygray}{-}
&\cellcolor{mygray}{-}
&0.6425
&0.7441
&\cellcolor{mygray}{-}
&\cellcolor{mygray}{-}
&\cellcolor{mygray}{-}
&\cellcolor{mygray}{-}
&\cellcolor{mygray}{-}
&\cellcolor{mygray}{-}
\\
WeakPolyp~\cite{WeakPolyp}
&MICCAI’23
&PVTv2-B2~\cite{PVTv2}
&\cellcolor{mygray}{-}
&\cellcolor{mygray}{-}
&\cellcolor{mygray}{-}
&\cellcolor{mygray}{-}
&\cellcolor{mygray}{-}
&\cellcolor{mygray}{-}
&\cellcolor{mygray}{-}
&\cellcolor{mygray}{-}
&0.7490
&0.8066
&\cellcolor{mygray}{-}
&\cellcolor{mygray}{-}
&\cellcolor{mygray}{-}
&\cellcolor{mygray}{-}
&\cellcolor{mygray}{-}
&\cellcolor{mygray}{-}
\\
Inf-Net~\cite{Inf-Net}
&TMI’20
&Res2Net-50~\cite{Res2Net}
&\cellcolor{mygray}{-}
&\cellcolor{mygray}{-}
&\cellcolor{mygray}{-}
&\cellcolor{mygray}{-}
&\cellcolor{mygray}{-}
&\cellcolor{mygray}{-}
&\cellcolor{mygray}{-}
&\cellcolor{mygray}{-}
&\cellcolor{mygray}{-}
&\cellcolor{mygray}{-}
&0.4324
& 0.5285
&\cellcolor{mygray}{-}
&\cellcolor{mygray}{-}
&\cellcolor{mygray}{-}
&\cellcolor{mygray}{-}
\\
DECOR-Net~\cite{DECOR-Net}
&ISBI’23
&Customized Design
&\cellcolor{mygray}{-}
&\cellcolor{mygray}{-}
&\cellcolor{mygray}{-}
&\cellcolor{mygray}{-}
&\cellcolor{mygray}{-}
&\cellcolor{mygray}{-}
&\cellcolor{mygray}{-}
&\cellcolor{mygray}{-}
&\cellcolor{mygray}{-}
&\cellcolor{mygray}{-}
&0.4025
&0.6949
&\cellcolor{mygray}{-}
&\cellcolor{mygray}{-}
&\cellcolor{mygray}{-}
&\cellcolor{mygray}{-}
\\
AAU-net~\cite{AAU-net}
&TMI’22
&Customized Design
&\cellcolor{mygray}{-}
&\cellcolor{mygray}{-}
&\cellcolor{mygray}{-}
&\cellcolor{mygray}{-}
&\cellcolor{mygray}{-}
&\cellcolor{mygray}{-}
&\cellcolor{mygray}{-}
&\cellcolor{mygray}{-}
&\cellcolor{mygray}{-}
&\cellcolor{mygray}{-}
&\cellcolor{mygray}{-}
&\cellcolor{mygray}{-}
&0.4745
&0.6515
&\cellcolor{mygray}{-}
&\cellcolor{mygray}{-}
\\
CMUNet~\cite{CMUNet}
&ISBI’23
&Customized Design
&\cellcolor{mygray}{-}
&\cellcolor{mygray}{-}
&\cellcolor{mygray}{-}
&\cellcolor{mygray}{-}
&\cellcolor{mygray}{-}
&\cellcolor{mygray}{-}
&\cellcolor{mygray}{-}
&\cellcolor{mygray}{-}
&\cellcolor{mygray}{-}
&\cellcolor{mygray}{-}
&\cellcolor{mygray}{-}
&\cellcolor{mygray}{-}
&0.5452
&0.8302
&\cellcolor{mygray}{-}
&\cellcolor{mygray}{-}
\\
MALUNet~\cite{MALUNet}
&BIBM’22
&Customized Design
&\cellcolor{mygray}{-}
&\cellcolor{mygray}{-}
&\cellcolor{mygray}{-}
&\cellcolor{mygray}{-}
&\cellcolor{mygray}{-}
&\cellcolor{mygray}{-}
&\cellcolor{mygray}{-}
&\cellcolor{mygray}{-}
&\cellcolor{mygray}{-}
&\cellcolor{mygray}{-}
&\cellcolor{mygray}{-}
&\cellcolor{mygray}{-}
&\cellcolor{mygray}{-}
&\cellcolor{mygray}{-}
&0.8632 
& 0.8537
\\
EGE-UNet~\cite{EGE-UNet}
&MICCAI’23
&Customized Design
&\cellcolor{mygray}{-}
&\cellcolor{mygray}{-}
&\cellcolor{mygray}{-}
&\cellcolor{mygray}{-}
&\cellcolor{mygray}{-}
&\cellcolor{mygray}{-}
&\cellcolor{mygray}{-}
&\cellcolor{mygray}{-}
&\cellcolor{mygray}{-}
&\cellcolor{mygray}{-}
&\cellcolor{mygray}{-}
&\cellcolor{mygray}{-}
&\cellcolor{mygray}{-}
&\cellcolor{mygray}{-}
&0.8588
&0.8498
\\
  \hline
    \multicolumn{19}{c}{{\large\textbf{Unified Models
}}}
    \\
    \hline
EVP~\cite{EVP}
&CVPR’23
&MiT-B4~\cite{SegFormer}
&0.8431
&0.9007
&0.7262
&0.8346
&0.0481
&0.0312
&\cellcolor{mygray}{-}
&\cellcolor{mygray}{-}
&\cellcolor{mygray}{-}
&\cellcolor{mygray}{-}
&\cellcolor{mygray}{-}
&\cellcolor{mygray}{-}
&\cellcolor{mygray}{-}
&\cellcolor{mygray}{-}
&\cellcolor{mygray}{-}
&\cellcolor{mygray}{-}
\\
UniverSeg~\cite{UniverSeg}
&ICCV’23
&ResNet-101~\cite{ResNet}
&\cellcolor{mygray}{-}
&\cellcolor{mygray}{-}
&\cellcolor{mygray}{-}
&\cellcolor{mygray}{-}
&\cellcolor{mygray}{-}
&\cellcolor{mygray}{-}
&\cellcolor{mygray}{-}
&\cellcolor{mygray}{-}
&0.5525
&0.2610
&0.6726
&0.3676
&0.7749
&0.5998
&0.7605
&0.7082
\\
SegGPT~\cite{SegGPT}
&ICCV’23
&ViT-L~\cite{ViT}
&0.3874
&0.6283
&0.4041
&0.6529
&0.4640
&0.2041
&0.4631
&0.3064
&0.5677
&0.7074
&0.1309
&0.5533
&0.3455
&0.6033
&0.4803
&0.4402
\\
\rowcolor{mygray1}
Spider
&-
&ViT-B~\cite{ViT}
&0.8679
&0.9074
&0.7532
&0.8505
&0.0440
&0.0308
&0.0680
&0.0550
&0.8038
&0.8540
&0.6913
&0.8118
&0.8254
&0.8607
&0.8948
&0.8758
\\
\rowcolor{mygray1}
Spider
&-
&ViT-L~\cite{ViT}
&0.8704
&0.9102
&0.7720
&0.8615
&0.0429
&0.0284
&0.0632
&0.0545
&0.7965
&0.8554 
&0.6915
&0.8128
&0.8298
&0.8599
&0.8954
&0.8743
\\
\rowcolor{mygray1}
Spider
&-
&Swin-B~\cite{Swin}
&0.8688
&0.9086
&0.7562
&0.8527
&0.0438
&0.0302
&0.0673
&0.0547
&0.8033
&0.8561
&0.6927
&0.8121
&0.8297
&0.8614
&\color{reda}\textbf{0.8968}
&\color{reda}\textbf{0.8767}
\\
\rowcolor{mygray1}
Spider
&-
&Swin-L~\cite{Swin}
&0.8729
&0.9109
&0.7731
&0.8620
&0.0423
&0.0271
&\color{reda}\textbf{0.0628}
&0.0539
&0.7975
&0.8550 
&0.6923
&0.8121
&0.8310
&0.8609
&0.8961
&0.8757
\\
\rowcolor{mygray1}
Spider
&-
&ConvNeXt-B~\cite{ConvNeXt}
&0.8732
&0.9109
&0.7779
&0.8625
&0.0444
&0.0272
&0.0632
&\color{reda}\textbf{0.0522}
&0.8211
&0.8655
&0.6925
&0.8106
&0.8352
&0.8632
&0.8949
&0.8733
\\
\rowcolor{mygray1}
Spider
&-
&ConvNeXt-L~\cite{ConvNeXt}
&\color{reda}\textbf{0.8821}
&\color{reda}\textbf{0.9158}
&\color{reda}\textbf{0.7893}
&\color{reda}\textbf{0.8674}
&\color{reda}\textbf{0.0396}
&\color{reda}\textbf{0.0265}
&{0.0636}
&{0.0554}
&\color{reda}\textbf{0.8243}
&\color{reda}\textbf{0.8664}
&\color{reda}\textbf{0.6956}
&\color{reda}\textbf{0.8127}
&\color{reda}\textbf{0.8376}
&\color{reda}\textbf{0.8655}
&0.8943
&0.8735
    \\

\bottomrule[2pt]
\end{tabular}
	}
	\label{tab:comparison}
\end{table*}

\begin{figure*}[!t]
	\centering
	\includegraphics[width=1\linewidth]{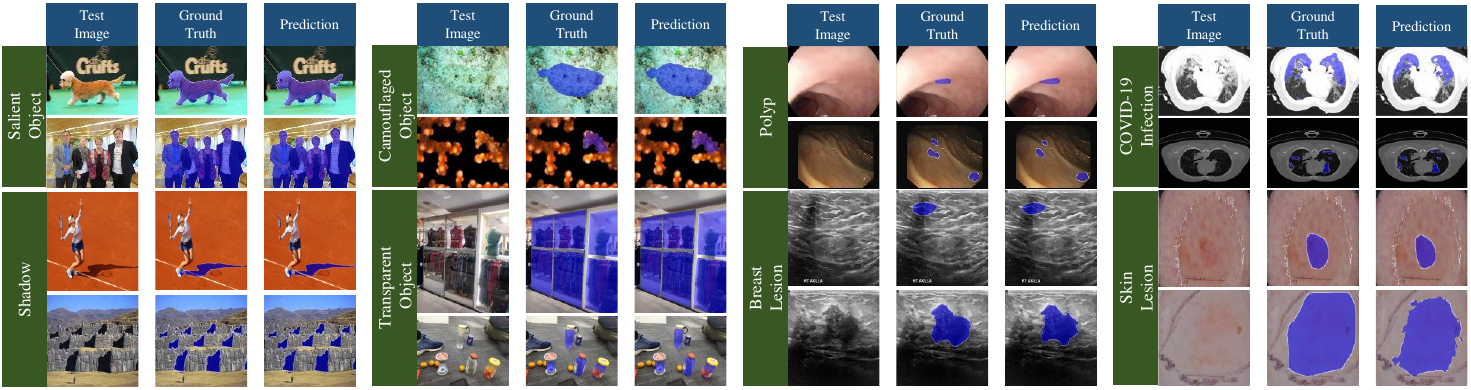}
	\vspace{-5mm}
	\caption{Some visual examples of Spider. Best viewed on screen.}
	\label{fig:visual_results_alltask}
	\vspace{-5mm}
\end{figure*}

\section{Experiments}

\subsection{Datasets and Metrics}
The dataset information is shown in~\cref{tab:Datasets_survey}. 
We follow the training settings of recent state-of-the-art methods in these tasks and merge all training samples  together as our training set. 
For evaluation, we introduce some widely used metrics, including
weighted F-measure~\cite{Fwb} ($F_{\beta}^{\omega}$) and S-measure~\cite{S-m} ($S_m$) for SOD and COD,
BER~\cite{BER} and MAE for SD and TOS,
and mIoU and mDice for the medical segmentation tasks.

\subsection{Implementation Details}
We follow many visual unified models~\cite{SegGPT,UNINEXT,Unicorn,UniHCP,SAM,Painter} to adopt a strong backbone as the encoder for covering the rich information from different large-scale datasets, which has become a consensus in current unified modeling field. 
In this work, we separately adopt the Transformer-based ViT~\cite{ViT}, Swin~\cite{Swin} and CNN-based ConvNeXt~\cite{ConvNeXt} as the visual encoder to demonstrate the performance of the proposed Spider.
All the experiments are implemented on the 8 Tesla A100 GPU for training $50$ epochs.
The input resolutions of images are resized to $384\times384$.
For each task, the mini-batch sizes of the input and prompt are set to $4$ and $12$, respectively.
We adopt some basic image augmentation techniques to avoid overfitting, including random flipping, rotating and border clipping. 
The Adam~\cite{Adam} optimizer scheduled by ``step'' with initial learning rate of $0.0001$, decay size of $30$ and decay rate of $0.9$ is introduced to update model parameters.

\begin{figure}[t]
	\centering
	\includegraphics[width=\linewidth]{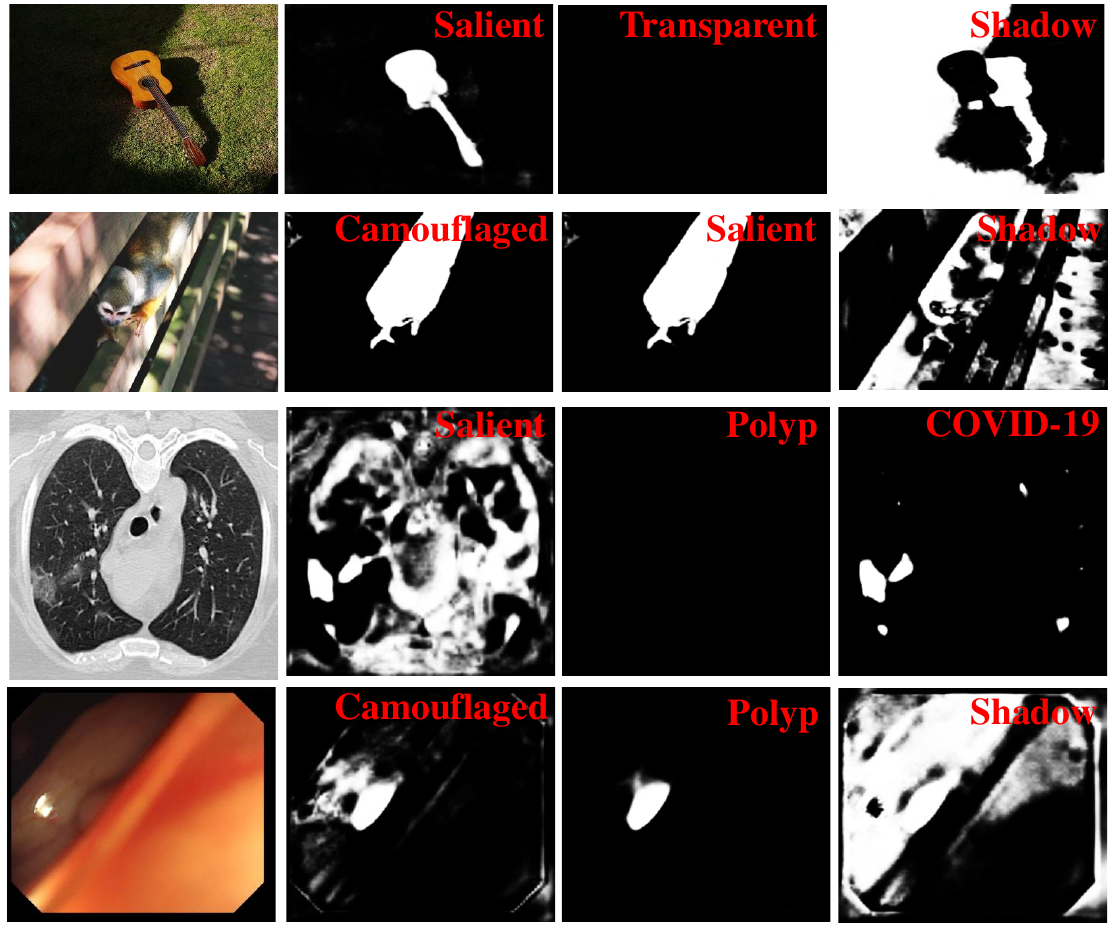}
	\vspace{-5mm}
	\caption{
		Visual predictions of multiple concepts in an image.
	}
	\label{fig:viusal_results1}
	\vspace{-5mm}
\end{figure}
\subsection{Evaluation}

\noindent\textbf{Quantitative Results.}
We compare the Spider with recent state-of-the-art specialized models and unified models as shown in~\cref{tab:comparison}.
It can be seen that Spider achieves dominant performance on all the tasks and performs better than each specialized model.  
In particular, it outperform other competitors by more than 30\% on the TOS, CLI and BLS tasks.
For the unified methods, EVP~\cite{EVP} only unifies three tasks and relays on three sets of adaptor parameters. UniverSeg~\cite{UniverSeg} only focuses on medical image segmentation tasks.
SegGPT~\cite{SegGPT} and Spider are able to accomplish all tasks with a single set of parameters. 
Limited by the prompt strategy based on a single image-mask pair, SegGPT cannot show the generalization ability across these tasks involving context-dependent concepts, even if it has been trained on more than 250,000 diverse images.

\noindent\textbf{Qualitative Results.}
We show some visual results in~\cref{fig:visual_results_alltask}. 
The detailed group prompts for all tasks and qualitative comparisons with other methods can be found in~\cref{sec:vis_cgp} and~\cref{sec:qualitative_comp}. 
In addition, Spider has multi-concept understanding ability as shown in~\cref{fig:viusal_results1}. There are some insightful phenomena.
For the monkey (see the 2$^{nd}$ row), Spider predicts  salient and camouflaged object segmentation map at the same time.
According to the intuitive response of human vision system, zooming out makes the monkey hidden in the surrounding environment, but zooming in makes it slightly stand out. 
Therefore, the concepts of saliency and camouflage may coexist and even sometimes are manifested in the same object.
It is also in line with the research motivation of salient and camouflaged object ranking~\cite{salient_ranking1,salient_ranking2,camo_ranking1,camo_ranking2}.
For the colon image (see the 4$^{th}$ row), 
we try to elicit the concepts of camouflage, polyp, and shadows.
Polyp lesions are usually hidden on the surface of the colon, our camouflaged object prediction effectively perceives polyp, which illustrates that the COD data is possibly beneficial to polyp segmentation.
Utilizing a large amount of natural scene data to improve medical lesion segmentation will promote the semi-supervised learning research in medical image field.
Finally, we also provide a good shadow detection for the colon image, which reveals a potential application of colonoscope shadow removal for improving the lesion visualization of the medical equipment.

\begin{table*}[!t]
	\centering
	\caption{Ablation studies on the eight tasks. All models adopt the ConvNeXt-B as the backbone.
	}
	\resizebox{\linewidth}{!}{
		\setlength\tabcolsep{2pt}
		\renewcommand\arraystretch{1.2}
		\begin{tabular}{l|cc|cc|cc|cc|cc|cc|cc|cc}
\toprule[2pt]

   &   \multicolumn{2}{c|}{\large\textbf{{Salient}}}
   & \multicolumn{2}{c|}{\large\textbf{{Camouflaged
}}}
   & \multicolumn{2}{c|}{\large\textbf{{Shadow}}}& \multicolumn{2}{c|}{\large\textbf{{Transparent}}}
   & \multicolumn{2}{c|}{\large\textbf{{Polyp}}}
      & \multicolumn{2}{c|}{\large\textbf{{COVID-19}}}
    & \multicolumn{2}{c|}{\large\textbf{{Breast}}}
      & \multicolumn{2}{c}{\large\textbf{{Skin}}}
      \\
   & \multicolumn{2}{c|}{{{DUTS}}}
   & \multicolumn{2}{c|}{{{COD10K
}}}
   & \multicolumn{2}{c|}{{{SBU}}}& \multicolumn{2}{c|}{{{Trans10K}}}
   & \multicolumn{2}{c|}{{{5 datasets}}}
      & \multicolumn{2}{c|}{{{COVID-19 CT}}}
    & \multicolumn{2}{c|}{{{BUSI}}}
      & \multicolumn{2}{c}{{{ISIC2018}}}
\\
     Method
     &F$_{\beta}^{\omega} \uparrow$ & S$_m \uparrow$
   &F$_{\beta}^{\omega} \uparrow$ & S$_m \uparrow$  &BER $\downarrow$ &MAE $\downarrow$  &BER $\downarrow$ &MAE $\downarrow$ &mDice $\uparrow$ &mIoU $\uparrow$ 
 &mDice $\uparrow$ &mIoU $\uparrow$ 
 &mDice $\uparrow$ &mIoU $\uparrow$ 
 &mDice $\uparrow$ &mIoU $\uparrow$ 
     \\
    \hline
    \multicolumn{17}{c}{{\large\textbf{(a) Joint Training vs. Separate Training 
}}}
    \\
    \hline
Separate Training
 &0.8593
 &0.9012
 &0.7543
 &0.8544
 &0.0476
 &0.0293
 &0.0673
 &0.0576
 &0.7786
 &0.8267
 &0.6367
 &0.7388
 &0.7747
 &0.7826
 &0.8548
 &0.8216
    \\
    \rowcolor{mygray1}
Joint Training
&0.8732
&0.9109
&0.7779
&0.8625
&0.0444
&0.0272
& 0.0632
& 0.0522
&0.8211
&0.8655
&0.6925
&0.8106
&0.8352
&0.8632
&0.8949
&0.8733
\\
    \hline
    \multicolumn{17}{c}{{\large\textbf{(b) Concept Filter
}}}
    \\
    \hline
UNet
&0.6253
&0.6345
&0.5346
&0.6038
&0.1382
&0.0846
&0.1426
&0.1135
&0.6144
&0.6532
&0.3672
&0.4112
&0.4378
&0.4782
&0.4889
&0.5023
\\
+ Image-Group Prompts
&0.7843
&0.8346
&0.7055
&0.7887
&0.0534
&0.0332
&0.0778
&0.0685
&0.7230
&0.7564
&0.5732
&0.7038
&0.7301
&0.7901
&0.7903
&0.7888
\\
+ Mask-Group Prompts (Foreground)
&0.8422
&0.8907
&0.7523
&0.8302
&0.0473
&0.0298
&0.0674
&0.0581
&0.7809
&0.8388
&0.6509
&0.7631
&0.7746
&0.8316
&0.8573
&0.8501
\\
\rowcolor{mygray1}
+ Mask-Group Prompts (Background)
&0.8732
&0.9109
&0.7779
&0.8625
&0.0444
&0.0272
&0.0632
&0.0522
&0.8211
&0.8655
&0.6925
&0.8106
&0.8352
&0.8632
&0.8949
&0.8733
\\
Concept Filter $\rightarrow$ Addition Fusion
&0.6534
&0.6497
&0.5742
&0.6313
&0.1108
&0.0769
&0.1235
&0.1096
&0.6532
&0.6742
&0.3809
&0.4216
&0.5464
&0.5589
&0.5160
&0.5436
\\
    \hline
    \multicolumn{17}{c}{{\large\textbf{(c) Training Strategies in the Unified Framework
}}}
    \\
    \hline
Random FP - Unify BP 
&0.8608
&0.8998
&0.7562
&0.8573
&0.0453
&0.0280
&0.0655
&0.0547
&0.8102
&0.8607
&0.6340
&0.7538
&0.8008
&0.8212
&0.8778
&0.8645
\\
Balance FP - Separate BP
&0.8422
&0.8831
&0.7383
&0.8288
&0.0490
&0.0299
&0.0682
&0.0566
&0.7979
&0.8425
&0.6388
&0.7612
&0.8046
&0.8308
&0.8508
&0.8477
\\
\rowcolor{mygray1}
Balance FP - Unify BP
&0.8732
&0.9109
&0.7779
&0.8625
&0.0444
&0.0272
&0.0632
&0.0522
&0.8211
&0.8655
&0.6925
&0.8106
&0.8352
&0.8632
&0.8949
&0.8733
\\
    \hline
    \multicolumn{17}{c}{{\large\textbf{(d) Number and Selection of Prompts
}}}
    \\
    \hline
Random Selection (G = 1)
&0.7038
&0.7134
&0.5732
&0.6789
&0.0789
&0.0520
&0.1136
&0.0971
&0.6533
&0.7011
&0.5620
&0.6844
&0.6135
&0.6346
&0.7421
&0.7116
\\
Random Selection (G = 4)
&0.8091
&0.8448
&0.6912
&0.7802
&0.0532
&0.0340
&0.0707
&0.0625
&0.7341
&0.7788
&0.6432
&0.7677
&0.7790
&0.8100
&0.8108
&0.7979
\\
Random Selection (G = 12)
&0.8348
&0.8815
&0.7298
&0.8064
&0.0488
&0.0310
&0.0685
&0.0574
&0.7736
&0.8164
&0.6527
&0.7809
&0.7977
&0.8209
&0.8402
&0.8316
\\
Random Selection (G = 64)
&0.8723
&0.9101
&0.7762
&0.8602
&0.0444
&0.0272
&0.0634
&0.0525
&0.8202
&0.8648
&0.6910
&0.8100
&0.8345
&0.8630
&0.8942
&0.8730
\\
\rowcolor{mygray1}
Clustering Selection (G = 64)
&0.8732
&0.9109
&0.7779
&0.8625
&0.0444
&0.0272
&0.0632
&0.0522
&0.8211
&0.8655
&0.6925
&0.8106
&0.8352
&0.8632
&0.8949
&0.8733
\\
\bottomrule[2pt]
    \end{tabular}
	}
	\vspace{-5mm}
	
	\label{tab:ablation_study}
\end{table*}

\begin{figure}[t]
	\centering
	\includegraphics[width=\linewidth]{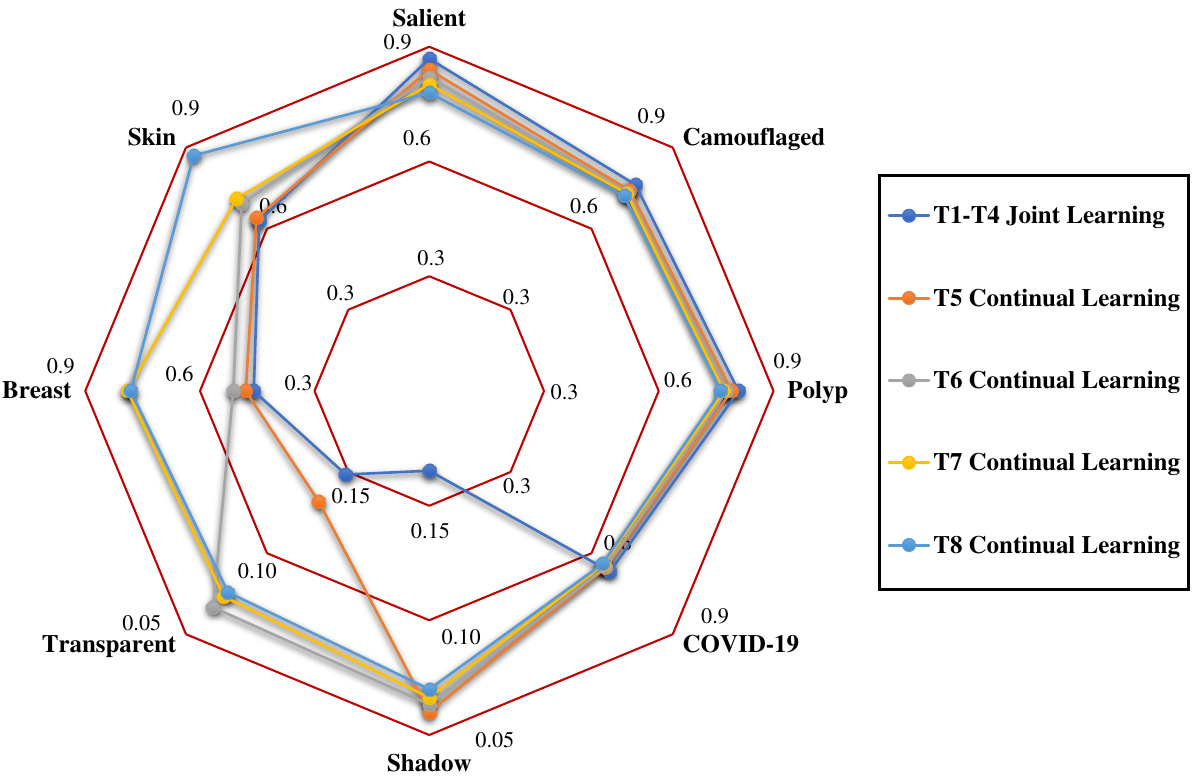}
	\vspace{-5mm}
	\caption{Performance of continuous learning on new tasks.
		Spider is firstly jointly pretrained on the T1-T4 data, and then continuously fine-tunes it from T5 to T8.
		T1: Salient Object Detection.
		T2: Camouflaged Object Detection.
		T3: Colon Polyp Segmentation.
		T4: COVID-19 Lung Infection.
		T5: Shadow Detection.
		T6: Transparent Object Segmentation.
		T7: Breast Lesion Segmentation.
		T8: Skin Lesion Segmentation.
	}
	\label{fig:continual_learning}
	\vspace{-5mm}
\end{figure}

\subsection{Ablation Study}
In~\cref{tab:ablation_study}, all models are based on the ConvNeXt-B~\cite{ConvNeXt} backbone.

\noindent\textbf{Joint Training vs. Separate Training.}
We train each model separately on each task with the same number of iterations and architectures as done in joint training. 
We can observe that the jointly trained models consistently outperform the separately trained ones on all tasks. 
This indicates that our framework with 100\% shared parameters can assimilate rich cross-domain knowledge and well function in specific tasks with the help of the image-mask group prompts.

\noindent\textbf{Concept Filter.}
Our baseline is the general UNet~\cite{UNet} structure without any specified design. 
The concept filter aims to help the baseline improve scene understanding and task discrimination. 
We step by step verify the prompts that drive the concept filter.
First, image-group prompts have the basic ability to find task commonality from image group, which significantly improves the performance over UNet on all tasks by more than 25\%.
Then, the foreground features are used as the query of transformer to directly establish the contrast relationship between the object query and the whole image.
In this way, `` + Mask-Group Prompts (Foreground) '' achieves similar performance with the separate training model.
Next, the background features are introduced to highlight the importance of the surroundings for concept expression, which achieves  40\% performance gain compared to the baseline. 
Finally, we replace the concept filter with the element-wise addition fusion  (keeping similar number of parameters) to show the advantages of the proposed high-level concept matching mechanism.

\noindent\textbf{Training Strategies in the Unified Framework.}
We conduct the experiments in terms of forward and back propagation, including random data partition and separate gradient update for each task. 
We can see that ``Balance FP - Unify BP'' performs the best,
which suggests that when training a unified model, all task data should be treated as a whole and each part is equally important.
Belittling any one of them will produce negative effect to other tasks.

\noindent\textbf{Number and Selection of Prompts.}
We evaluate the impact of different number of random prompts in the inference phase.
It can be seen that the overall performance is the worst when $G = 1$.
As the number increases, the performance is gradually elevated and stabilizes when $G = 64$.
Moreover, we select 64 pairs of samples as the group prompts by clustering training data for each task.
It can be seen that ``Clustering Selection ($G = 64$)'' has almost the same performance as ``Random Selection''.
Thus, the strategy of random selection during training indeed makes Spider robust against different group prompts when testing.  More experiments of prompts during training and testing phases can be found in the~\cref{sec:number_prompts}.

\noindent\textbf{Continuous Learning \& Potential for Unseen Tasks.}
\cref{fig:continual_learning} shows the ability of continuous learning of Spider.
First, we jointly train Spider on the four tasks including SOD, COD, CPS, and CLI, to ensure the basic general segmentation capability. 
Then, the continuous learning is performed on the additional training sets from T5 to T8, where we only fine-tune the last layer of the decoder and the concept filter. 
The minimal number of trainable parameters ($<1\%$) drastically accelerates the training process, and alleviates the catastrophic forgetting~\cite{cl1,cl2}.
It can be seen that Spider's performance on new tasks is significantly improved, 
while there is only a negligible performance degradation of no more than 5\% on the old tasks.
Besides, the performance for SLS is over 0.6 mIoU even if the model is only trained on the T1-T4 data.
With the increasing of data scale and diversity, the performance is steadily improved in some unseen tasks. More relevant analysis are be found in the~\cref{sec:IZSL}.

\section{Conclusion}

We propose Spider, a universal context-dependent concept understanding
model, to unify eight segmentation tasks with the proposed group prompt paradigm. 
Extensive experiments demonstrate the superior performance of the proposed Spider on twelve challenging benchmarks using a single set of parameters. 
Spider can serve as a solid baseline within the unified cross-domain research. 
In the future, we will expand Spider to more context-dependent concept understanding tasks, such as industrial defect detection, inharmonious region localization, and defocus blur detection.
We are also working on introducing image editing tasks into the Spider framework, which can simultaneously complete more interesting applications such as shadow detection and removal, salient object detection and camouflageization, inharmonious region localization and harmonization.

\section*{Acknowledgements}
We thank all the reviewers for their feedbacks through out the review cycles of the manuscript. 
We are very grateful to Dr. Xinlong Wang for his constructive suggestions on this work. 
This work was supported by the National Natural Science Foundation of China under Grant 62276046 and by Dalian Science and Technology Innovation Foundation under Grant 2023JJ12GX015.

\section*{Impact Statement}
This paper presents work whose goal is to advance the field of 
Machine Learning. There are many potential societal consequences 
of our work, none which we feel must be specifically highlighted here.

\nocite{langley00}

\bibliography{example_paper}
\bibliographystyle{icml2024}

\newpage
\appendix
\onecolumn

\begin{table*}[t]
	\centering
	\caption{Performance stability when using low qualities of mask annotations of the prompts during inference.
	}
	\resizebox{\linewidth}{!}{
		\setlength\tabcolsep{2pt}
		\renewcommand\arraystretch{1.2}
		\begin{tabular}{c|cc|cc|cc|cc|cc|cc|cc|cc}
\toprule[2pt]

    &  \multicolumn{2}{c|}{\large\textbf{{Salient}}}
   & \multicolumn{2}{c|}{\large\textbf{{Camouflaged
}}}
   & \multicolumn{2}{c|}{\large\textbf{{Shadow}}}& \multicolumn{2}{c|}{\textbf{{Transparent}}}
   & \multicolumn{2}{c|}{\large\textbf{{Polyp}}}
      & \multicolumn{2}{c|}{\large\textbf{{COVID-19}}}
    & \multicolumn{2}{c|}{\large\textbf{{Breast}}}
      & \multicolumn{2}{c}{\large\textbf{{Skin}}}
      \\
  & \multicolumn{2}{c|}{{{DUTS}}}
   & \multicolumn{2}{c|}{{{COD10K
}}}
   & \multicolumn{2}{c|}{{{SBU}}}& \multicolumn{2}{c|}{{{Trans10K}}}
   & \multicolumn{2}{c|}{{{5 datasets}}}
      & \multicolumn{2}{c|}{{{COVID-19 CT}}}
    & \multicolumn{2}{c|}{{{BUSI}}}
      & \multicolumn{2}{c}{{{ISIC2018}}}
\\
    Mask Prompt 
     &F$_{\beta}^{\omega} \uparrow$ & S$_m \uparrow$
   &F$_{\beta}^{\omega} \uparrow$ & S$_m \uparrow$  &BER $\downarrow$ &MAE $\downarrow$  &BER $\downarrow$ &MAE $\downarrow$ &mDice $\uparrow$ &mIoU $\uparrow$ 
 &mDice $\uparrow$ &mIoU $\uparrow$ 
 &mDice $\uparrow$ &mIoU $\uparrow$ 
 &mDice $\uparrow$ &mIoU $\uparrow$ 
     \\
    \hline
Ground Truth
&0.8732
&0.9109
&0.7779
&0.8625
&0.0444
&0.0272
&0.0632
&0.0522
&0.8211
&0.8655
&0.6925
&0.8106
&0.8352
&0.8632
&0.8949
&0.8733
\\
Dilation: Kernel = $3 \times 3$
&0.8718
&0.9100
&0.7771
&0.8615
&0.0446
&0.0275
&0.0638
&0.0525
&0.8205
&0.8651
&0.6922
&0.8102
&0.8348
&0.8630
&0.8946
&0.8730
\\
Dilation: Kernel = $5 \times 5$
&0.8718
&0.9102
&0.7768
&0.8617
&0.0446
&0.0278
&0.0639
&0.0526
&0.8204
&0.8650
&0.6922
&0.8102
&0.8345
&0.8631
&0.8945
&0.8728
\\
Dilation: Kernel = $7 \times 7$
&0.8714
&0.9100
&0.7764
&0.8614
&0.0448
&0.0280
&0.0641
&0.0525
&0.8204
&0.8651
&0.6921
&0.8103
&0.8346
&0.8634
&0.8944
&0.8725
\\
Erosion: Kernel = $3 \times 3$
&0.8714
&0.9098
&0.7764
&0.8603
&0.0448
&0.0278
&0.0641
&0.0530
&0.8202
&0.8655
&0.6917
&0.8097
&0.8345
&0.8627
&0.8940
&0.8725
\\
Erosion: Kernel = $5 \times 5$
&0.8708
&0.9074
&0.7754
&0.8589
&0.0450
&0.0279
&0.0645
&0.0534
&0.8189
&0.8643
&0.6915
&0.8090
&0.8338
&0.8617
&0.8934
&0.8721
\\
Erosion: Kernel = $7 \times 7$
&0.8699
&0.9063
&0.7732
&0.8578
&0.0453
&0.0284
&0.0650
&0.0545
&0.8158
&0.8635
&0.6898
&0.8068
&0.8321
&0.8600
&0.8922
&0.8715
\\
\bottomrule[2pt]
    \end{tabular}
	}
	
	\label{tab:ablation_mask_acc}
\end{table*}
\section{Appendix}
\subsection{Advantages of the Proposed Concept Filter}\label{sec:advantages_CF}
\textit{\textbf{\uppercase\expandafter{\romannumeral1})}}  \textit{Robustness}: 
The model can use extra background features to regularize representation learning instead of relying solely on foreground features.
\textit{\textbf{\uppercase\expandafter{\romannumeral2})}}  \textit{Learning Efficiency}: Dividing features into foreground and background can make it easier for the model to learn important features. Since different features are represented in the weights and biases, this helps the model converge faster.
\textit{\textbf{\uppercase\expandafter{\romannumeral3})}}  \textit{Interpretability}: Splitting the mask-prompts into foreground and background parts not only helps researchers better understand how the model works, but also increases the model's credibility.
\textit{\textbf{\uppercase\expandafter{\romannumeral4})}}  \textit{Flexibility}: If there is no obvious background information in some images, the model can leverage the background feature generator to generate meaningful biases.
\textit{\textbf{\uppercase\expandafter{\romannumeral5})}}  \textit{Generalization Ability}: By utilizing foreground and background features separately, the model can better adapt to different data distributions. It can improve the generalization ability of the model, allowing it to  handle unseen images.
\textit{\textbf{\uppercase\expandafter{\romannumeral6})}}  \textit{High Tolerance Rate of Prompt Annotation}: Different from direct pix-level feature fusion, condensing prompt knowledge into high-level expression filters can reduce the model's requirements for mask annotation accuracy of prompts. As shown in \cref{tab:ablation_mask_acc}, our performance is stable when faced with varying degrees of dilation and erosion on the prompt mask.

\subsection{Definition of Different Context-dependent Image Segmentation Tasks} \label{sec:def_CIS}
\textit{\textbf{\uppercase\expandafter{\romannumeral1})}} \textit{Salient object detection} (SOD)~\cite{GateNet,MINet,DANet,HDFNet,SSLSOD,MMFT,CAVER,GateNetv2} is often associated with
\textit{\textbf{\uppercase\expandafter{\romannumeral2})}} \textit{camouflaged object detection} (COD)~\cite{ZoomNet,GateNetv2}.
The former aims at finding visually salient objects, while the latter focuses on hidden objects extremely similar to the surrounding backgrounds. 
\textit{\textbf{\uppercase\expandafter{\romannumeral3})}} \textit{Shadow detection} (SD) is an important research topic.
Because the shadow contains rich depth and geometry cues, shadow detection is often applied in image editing tasks, such as shadow removal~\cite{DAS_shadow_removal} and image synthesis~\cite{Arshadowgan}.
In addition, some important details of objects may be hidden in the shadow. It is very
necessary to understand the shadow. 
\textit{\textbf{\uppercase\expandafter{\romannumeral4})}} \textit{Transparent object segmentation} (TOS) is a challenging task due to the properties of reflection and refraction. 
It can assist indoor smart robots~\cite{Robotic_transparent} and outdoor unmanned logistics vehicles~\cite{vehicles_transparent} in controlling or avoiding transparent objects. 
In intelligent diagnosis, fully automatic image segmentation has become an important medical aid. 
Compared with the organs of fixed shape and appearance, the lesions have strong context-dependent concept, lesion segmentation is more challenging. 
\textit{\textbf{\uppercase\expandafter{\romannumeral5})}} \textit{Colon polyp segmentation} (CPS) \cite{PraNet_Polyp,MSNet_Polyp,WeakPolyp,LDNet_Polyp,EMS-Net_Polyp} identifies polyps of different sizes hidden on the surface of the intestinal wall. 
\textit{\textbf{\uppercase\expandafter{\romannumeral6})}} \textit{COVID-19 lung infection} (CLI)~\cite{DSC_COVID-19,BCS-Net_COVID-19,Anam-Net_COVID-19,wu2020jcs_COVID-19,Inf-Net,M2SNet} captures the infected area from a CT image containing other lung lesions and a large number of anatomical structures and tissue textures. 
\textit{\textbf{\uppercase\expandafter{\romannumeral7})}} \textit{Breast lesion segmentation} (BLS)~\cite{NU-net_breast,SKUNet_breast,chen2021domain_breast,AAU-net,CMUNet} needs to overcome the speckle noise in ultrasound images caused by the interaction of scattered sound waves and tissue structures, 
which reduces image contrast and blurs the lesion edge. 
\textit{\textbf{\uppercase\expandafter{\romannumeral8})}} \textit{Skin lesion segmentation} (SLS) \cite{dai2022ms_skin,lei2020skin_skin,wu2020automated_skin,MALUNet,EGE-UNet} aims to search dermatofibromas and epidermal cysts from dermoscopic images.

\subsection{Challenges of Context-dependent Concept Understanding}
\begin{figure*}[!t]
	\centering
	\includegraphics[width=\linewidth]{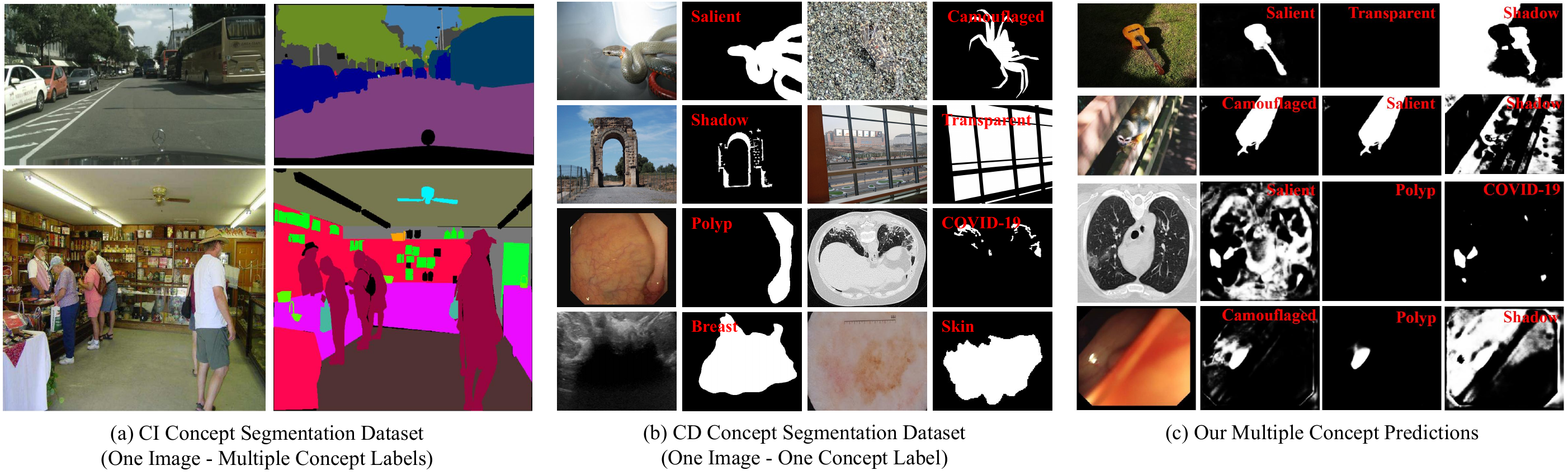}
	\caption{Context-independent concept segmentation datasets vs. Context-dependent concept segmentation datasets.}
	\label{fig:challenges_datasets}
	\vspace{-5mm}
\end{figure*}

For traditional semantic segmentation tasks, labeling data may be relatively easy. As shown in \cref{fig:challenges_datasets} (a), existing CI concept datasets, such as Cityscapes~\cite{Cityscapes} and ADE20K~\cite{ADE20K}, all have multiple concept annotations for a single image and do not overlap with each other. Current CI models can well distinguish different concepts. Existing CD concept datasets provide the annotates of single concept, as shown in \cref{fig:challenges_datasets} (b). 
Actually, multiple CD concepts often co-occur in an object. How to effectively depict the contrast between the foreground and background to highlight the characteristics of each concept is the key to achieving accurate segmentation. \cref{fig:challenges_datasets} (c) shows the multi-concept prediction capabilities obtained by Spider. In addition, concept-shift will produce in some moving objects, which puts higher requirements on the model's ability of concept understanding. In the future, we will study context-dependent concept understanding in video.

\begin{table*}[t]
	\centering
	\caption{Ablation experiments of the number of prompts in training and testing phases.
	}
	\resizebox{\linewidth}{!}{
		\setlength\tabcolsep{2pt}
		\renewcommand\arraystretch{1.2}
		\begin{tabular}{c|c|cc|cc|cc|cc|cc|cc|cc|cc}
\toprule[2pt]

  &  &  \multicolumn{2}{c|}{\large\textbf{{Salient}}}
   & \multicolumn{2}{c|}{\large\textbf{{Camouflaged
}}}
   & \multicolumn{2}{c|}{\large\textbf{{Shadow}}}& \multicolumn{2}{c|}{\textbf{{Transparent}}}
   & \multicolumn{2}{c|}{\large\textbf{{Polyp}}}
      & \multicolumn{2}{c|}{\large\textbf{{COVID-19}}}
    & \multicolumn{2}{c|}{\large\textbf{{Breast}}}
      & \multicolumn{2}{c}{\large\textbf{{Skin}}}
      \\
&   & \multicolumn{2}{c|}{{{DUTS}}}
   & \multicolumn{2}{c|}{{{COD10K
}}}
   & \multicolumn{2}{c|}{{{SBU}}}& \multicolumn{2}{c|}{{{Trans10K}}}
   & \multicolumn{2}{c|}{{{5 datasets}}}
      & \multicolumn{2}{c|}{{{COVID-19 CT}}}
    & \multicolumn{2}{c|}{{{BUSI}}}
      & \multicolumn{2}{c}{{{ISIC2018}}}
\\
    \#Train & \#Test 
     &F$_{\beta}^{\omega} \uparrow$ & S$_m \uparrow$
   &F$_{\beta}^{\omega} \uparrow$ & S$_m \uparrow$  &BER $\downarrow$ &MAE $\downarrow$  &BER $\downarrow$ &MAE $\downarrow$ &mDice $\uparrow$ &mIoU $\uparrow$ 
 &mDice $\uparrow$ &mIoU $\uparrow$ 
 &mDice $\uparrow$ &mIoU $\uparrow$ 
 &mDice $\uparrow$ &mIoU $\uparrow$ 
     \\
    \hline
1
&64
&0.7177
&0.7187
&0.6145
&0.6868
&0.0745
&0.0502
&0.0929
&0.0921
&0.6621
&0.7109
&0.5653
&0.6758
&0.6194
&0.6306
&0.7544
&0.7316
\\
4
&64
&0.8203
&0.8501
&0.7134
&0.7886
&0.0508
&0.0331
&0.0658
&0.0596
&0.7545
&0.7848
&0.6672
&0.7707
&0.7846
&0.8183
&0.8178
&0.8086
\\
\rowcolor{mygray1}
12
&64
&0.8732
&0.9109
&0.7779
&0.8625
&0.0444
&0.0272
&0.0632
&0.0522
&0.8211
&0.8655
&0.6925
&0.8106
&0.8352
&0.8632
&0.8949
&0.8733
\\
12
& 1
&0.7038
&0.7134
&0.5732
&0.6789
&0.0789
&0.0520
&0.1136
&0.0971
&0.6533
&0.7011
&0.5620
&0.6844
&0.6135
&0.6346
&0.7421
&0.7116
\\
12
& 4
&0.8091
&0.8448
&0.6912
&0.7802
&0.0532
&0.0340
&0.0707
&0.0625
&0.7341
&0.7788
&0.6432
&0.7677
&0.7790
&0.8100
&0.8108
&0.7979
\\
12
& 12
&0.8348
&0.8815
&0.7298
&0.8064
&0.0488
&0.0310
&0.0685
&0.0574
&0.7736
&0.8164
&0.6527
&0.7809
&0.7977
&0.8209
&0.8402
&0.8316
\\
12
& 64
&0.8723
&0.9101
&0.7762
&0.8602
&0.0444
&0.0272
&0.0634
&0.0525
&0.8202
&0.8648
&0.6910
&0.8100
&0.8345
&0.8630
&0.8942
&0.8730
\\
\rowcolor{mygray1}
1
& 1
&0.4674
&0.5389
&0.4375
&0.5745
&0.2346
&0.2541
&0.2406
&0.2720
&0.5935
&0.6345
&0.3784
&0.3990
&0.3046
&0.4589
&0.5846
&0.5038
\\
\bottomrule[2pt]
    \end{tabular}
	}
	
	\label{tab:ablation_study_training_number_prompts}
\end{table*}

\subsection{Applications of Context-dependent Concepts Understanding}
When multiple context-dependent concepts appear simultaneously in an image, the following applications will occur:
\textit{\textbf{\uppercase\expandafter{\romannumeral1})}} \textit{Human-computer interaction and virtual reality}: In human-computer interaction or virtual reality, salient objects in the user interface attract the user's attention. Shadow and transparency effects can be used to create more realistic 3D effects. Camouflage objects can be used to hide or show specific elements, while blurred backgrounds can help users focus on the main content.
\textit{\textbf{\uppercase\expandafter{\romannumeral2})}} \textit{Image Editing and Augmented Reality}: Transparent objects can be removed or blurred to make salient objects more prominent and improve the visual effect of the image.
\textit{\textbf{\uppercase\expandafter{\romannumeral3})}} \textit{Medical image analysis}: Spider can provide a good shadow detection for the colon image, which reveals a potential application of colonoscope shadow removal for improving the lesion visualization of the medical equipment.
\textit{\textbf{\uppercase\expandafter{\romannumeral4})}} \textit{Military reconnaissance and security inspection}: In the military or security field, salient objects in images may represent important military equipment or potential threats. Camouflaged objects may be used to hide true intentions or devices.
\textit{\textbf{\uppercase\expandafter{\romannumeral5})}} \textit{Autonomous driving}: In autonomous driving systems, it required to distinguish the salient objects such as vehicles, pedestrians and the camouflaged obstacles, transparent objects such as glass  to ensure the vehicle travels safely.
\textit{\textbf{\uppercase\expandafter{\romannumeral6})}} \textit{Environmental monitoring and urban planning}: In environmental monitoring and urban planning, by identifying multiple context-dependent concpets, we can understand the development of the city, environmental changes, and potential environment problems, etc., providing important basis for urban planning and environmental governance.

\subsection{Number of Prompts}
\label{sec:number_prompts}
In \cref{tab:ablation_study_training_number_prompts}, we thoroughly show the impact of different numbers of prompts during training and testing phases. The gap between the best choice (Train: 12, Test: 64) and the worst choice (Train: 1, Test: 1) demonstrates the necessity of group prompts for the model to understand the context-dependent concepts.

\subsection{Visualization of Clustered Group Prompts}
\label{sec:vis_cgp}
\begin{figure}[!t]
	\centering
	\includegraphics[width=\linewidth]{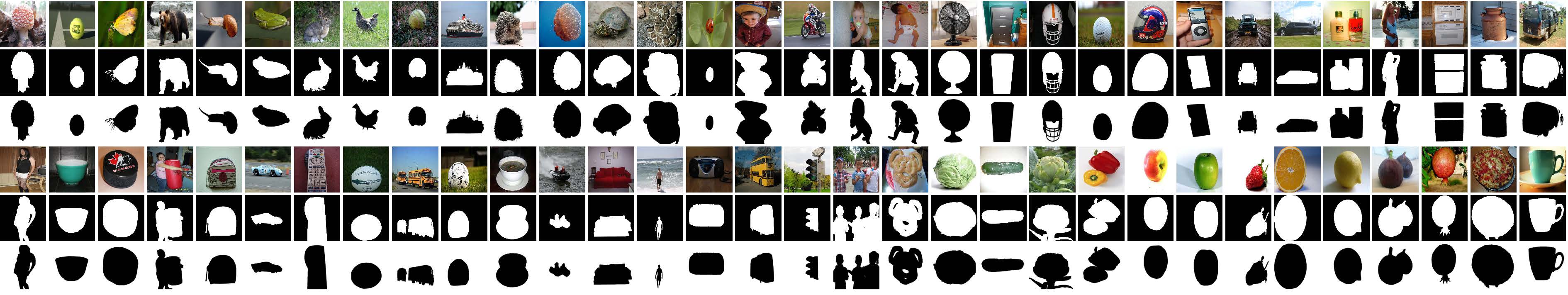}
	\caption{Visualization of clustered group prompts for Salient Object Detection.}
	\label{fig:prompt_sod}
\end{figure}

\begin{figure}[!t]
	\centering
	\includegraphics[width=\linewidth]{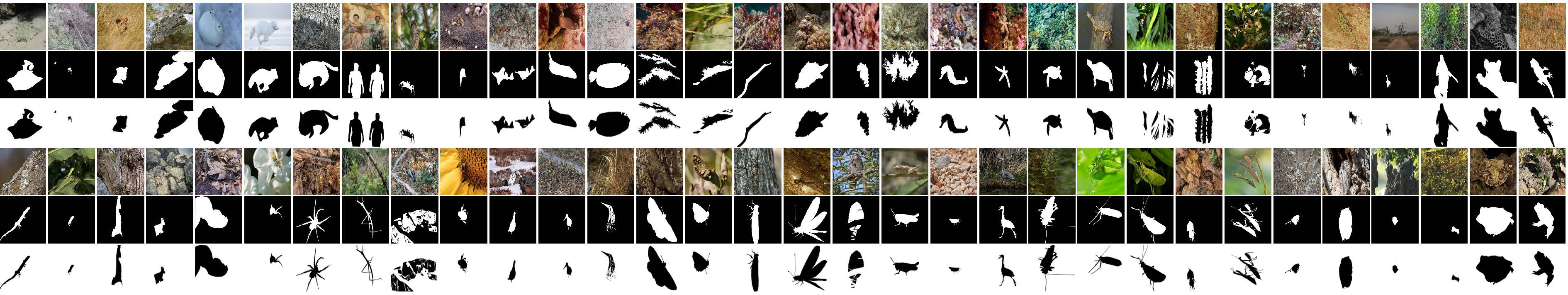}
	\caption{Visualization of clustered group prompts for Camouflaged Object Detection.}
	\label{fig:prompt_cod}
\end{figure}

\begin{figure}[!t]
	\centering
	\includegraphics[width=\linewidth]{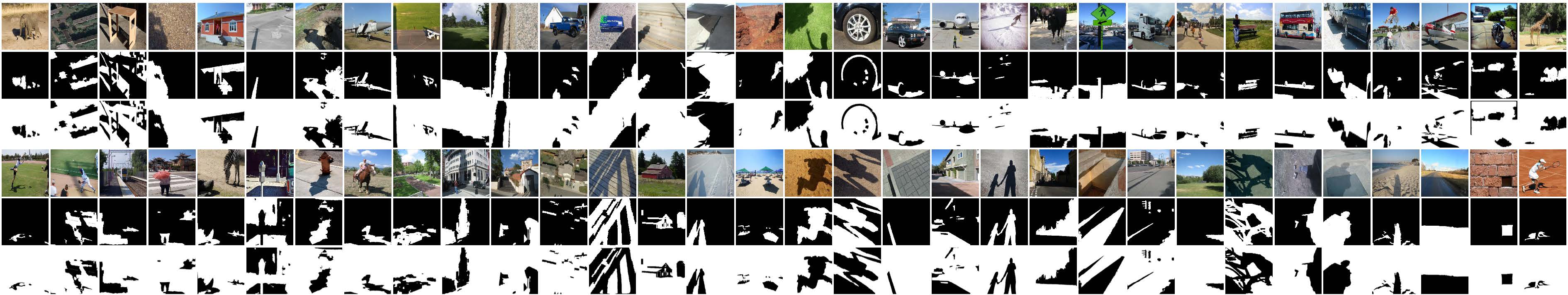}
	\caption{Visualization of clustered group prompts for Shadow Detection.}
	\label{fig:prompt_shadow}
\end{figure}

\begin{figure}[!t]
	\centering
	\includegraphics[width=\linewidth]{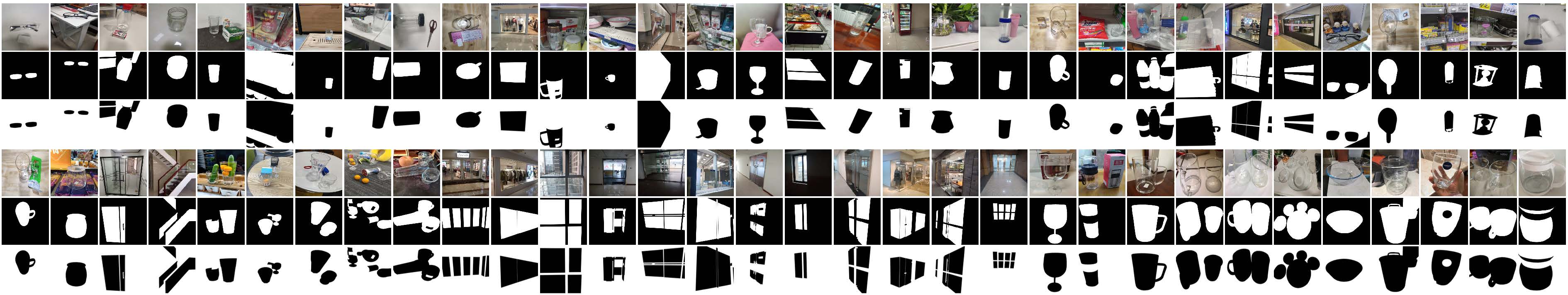}
	\caption{Visualization of clustered group prompts for Transparent Object Segmentation.}
	\label{fig:prompt_transparent}
\end{figure}

\begin{figure}[!t]
	\centering
	\includegraphics[width=\linewidth]{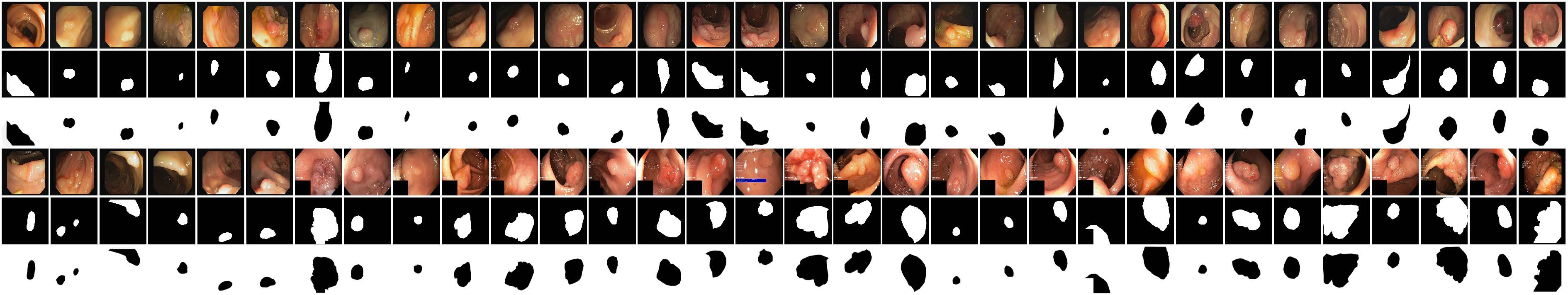}
	\caption{Visualization of clustered group prompts for Polyp Segmentation.}
	\label{fig:prompt_polyp}
\end{figure}

\begin{figure}[!t]
	\centering
	\includegraphics[width=\linewidth]{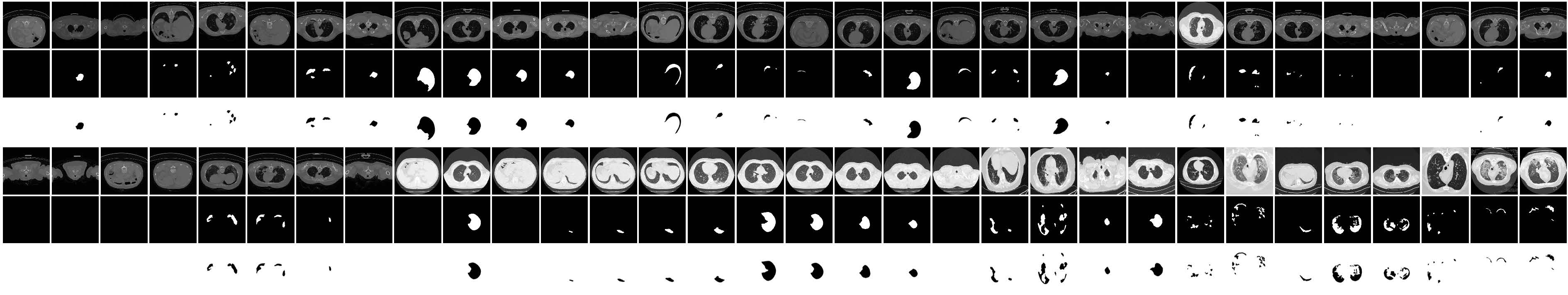}
	\caption{Visualization of clustered group prompts for COVID-19 Lung Infection.}
	\label{fig:prompt_covid}
\end{figure}

\begin{figure}[!t]
	\centering
	\includegraphics[width=\linewidth]{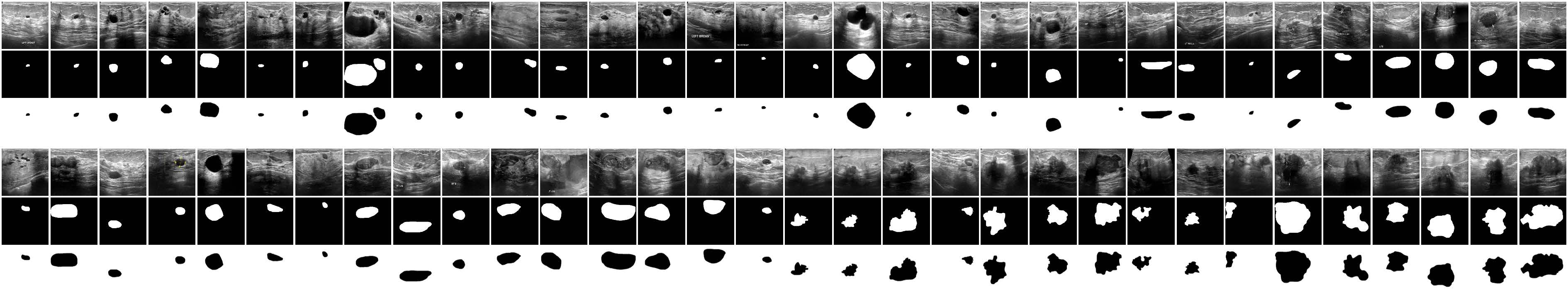}
	\caption{Visualization of clustered group prompts for Breast Lesion Segmentation.}
	\label{fig:prompt_breast}
\end{figure}

\begin{figure}[!t]
	\centering
	\includegraphics[width=\linewidth]{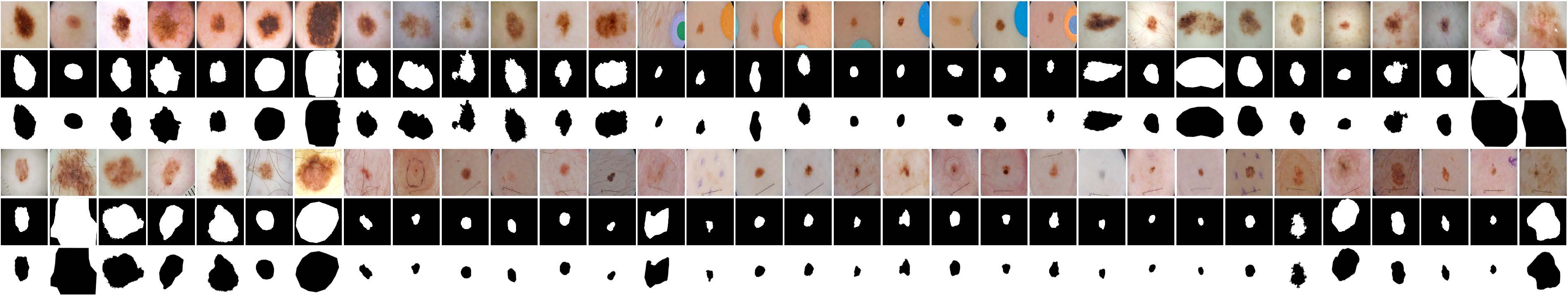}
	\caption{Visualization of clustered group prompts for Skin Lesion Segmentation.}
	\label{fig:prompt_skin}
\end{figure}

In \cref{fig:prompt_sod} - \cref{fig:prompt_skin}, we visualize the clustered group prompts used by each task during inference.

\subsection{Qualitative Comparisons}
\label{sec:qualitative_comp}
\begin{figure}[!t]
	\centering
	\includegraphics[width=\linewidth]{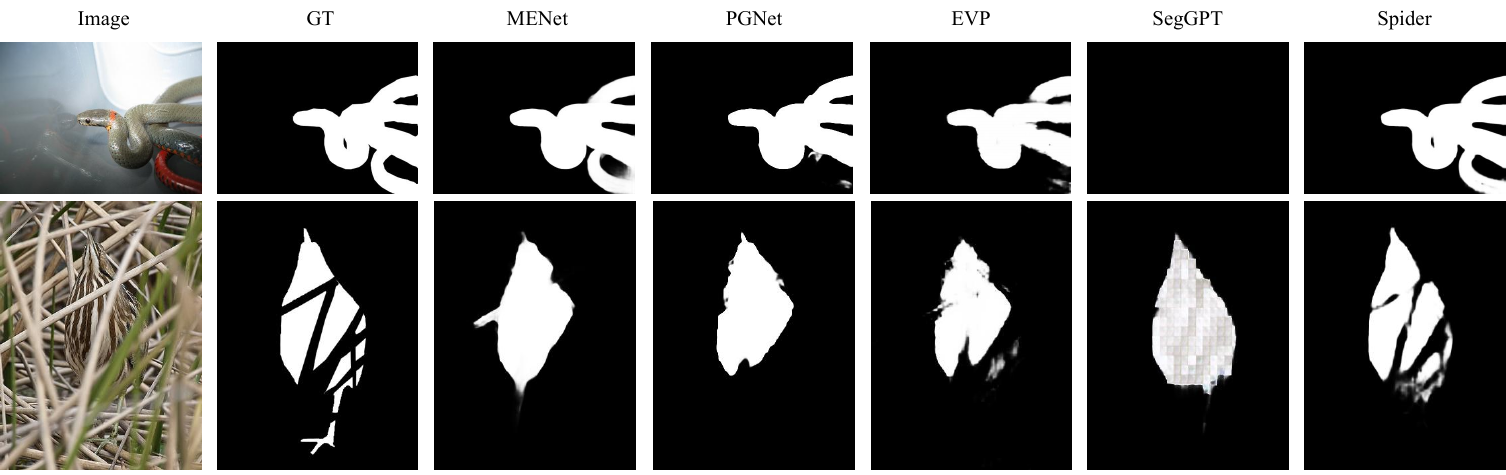}
	\caption{Qualitative comparisons with Salient Object Detection methods. MENet~\cite{MENet} and PGNet~\cite{PGNet} are the specialized models.}
	\label{fig:sod_comparison}
\end{figure}

\begin{figure}[!t]
	\centering
	\includegraphics[width=\linewidth]{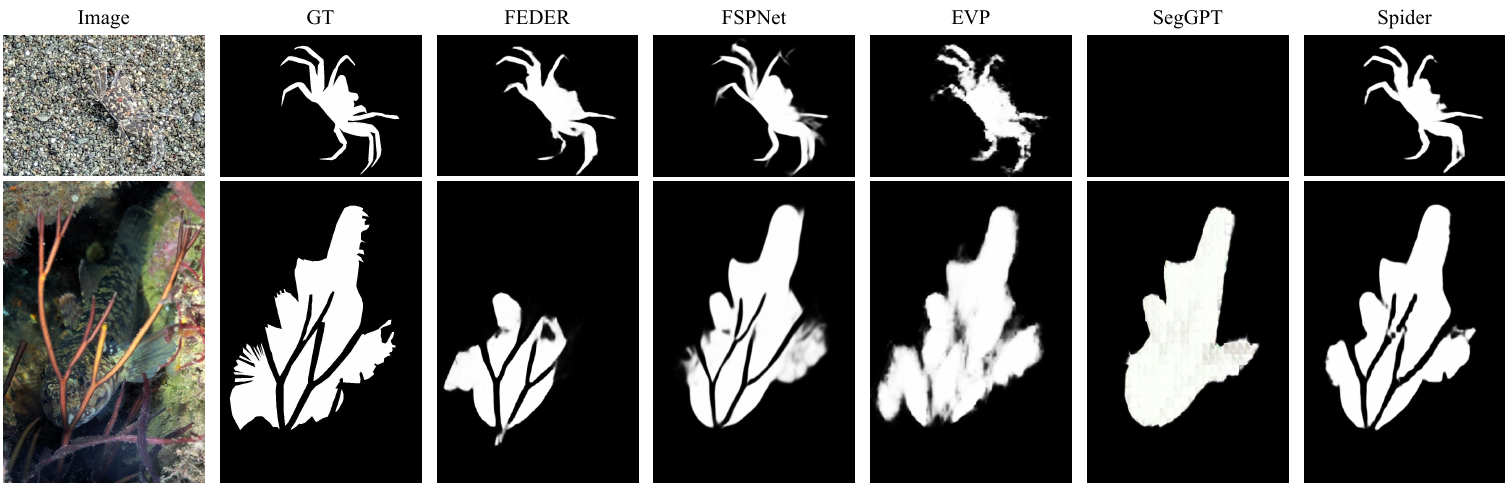}
	\caption{Qualitative comparisons with Camouflaged Object Detection methods. FEDER~\cite{FEDER} and FSPNet~\cite{FSPNet} are the specialized models.}
	\label{fig:cod_comparison}
\end{figure}

\begin{figure}[!t]
	\centering
	\includegraphics[width=\linewidth]{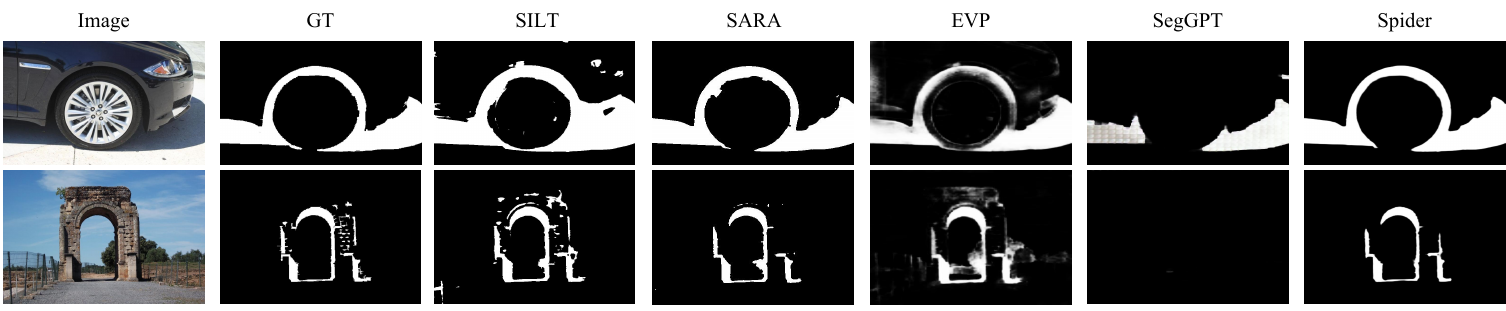}
	\caption{Qualitative comparisons with Shadow Detection methods. SILT~\cite{SILT} and SARA~\cite{SARA} are the specialized models.}
	\label{fig:shadow_comparison}
\end{figure}

\begin{figure}[!t]
	\centering
	\includegraphics[width=\linewidth]{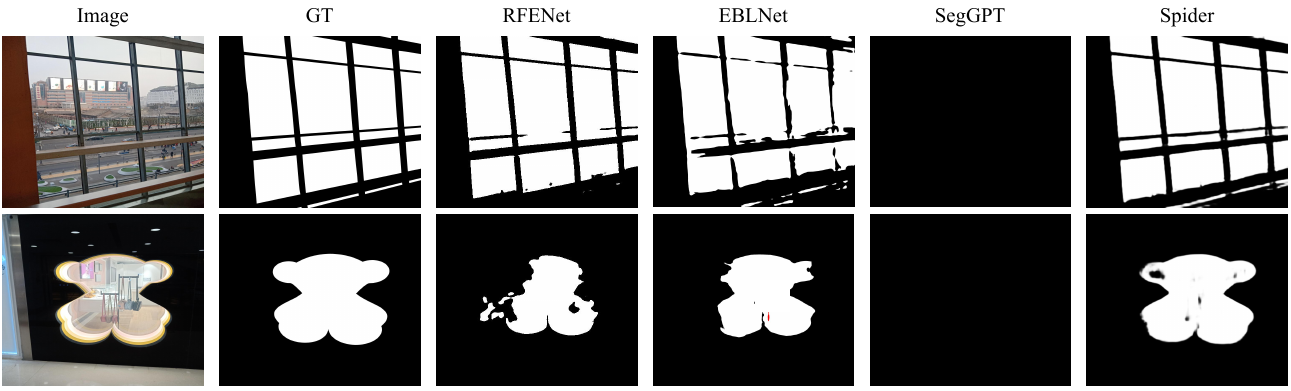}
	\caption{Qualitative comparisons with Transparent Object Segmentation methods. RFENet~\cite{RFENet} and EBLNet~\cite{EBLNet} are the specialized models.}
	\label{fig:transparent_comparison}
\end{figure}

\begin{figure}[!t]
	\centering
	\includegraphics[width=\linewidth]{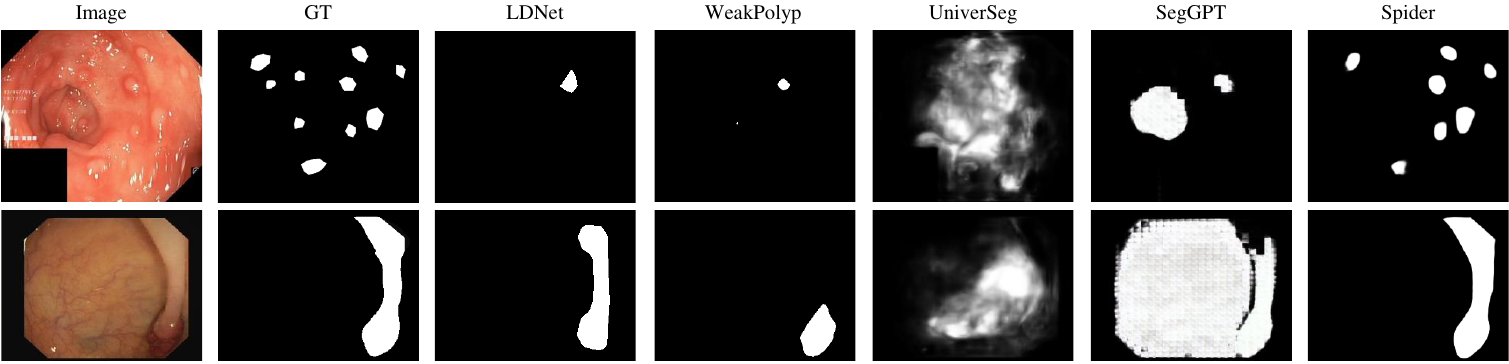}
	\caption{Qualitative comparisons with Polyp Segmentation methods. LDNet~\cite{LDNet_Polyp} and WeakPolyp~\cite{WeakPolyp} are the specialized models.}
	\label{fig:polyp_comparison}
\end{figure}

\begin{figure}[!t]
	\centering
	\includegraphics[width=\linewidth]{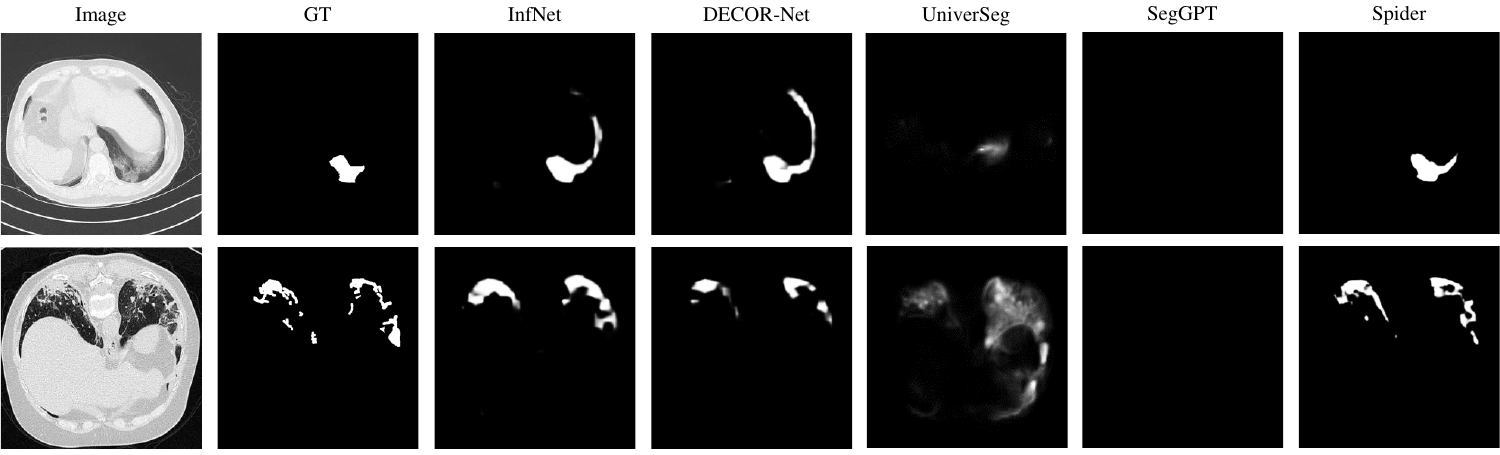}
	\caption{Qualitative comparisons with COVID-19 Lung Infection methods. InfNet~\cite{Inf-Net} and DECOR-Net~\cite{DECOR-Net} are the specialized models.}
	\label{fig:covid_comparison}
\end{figure}

\begin{figure}[!t]
	\centering
	\includegraphics[width=\linewidth]{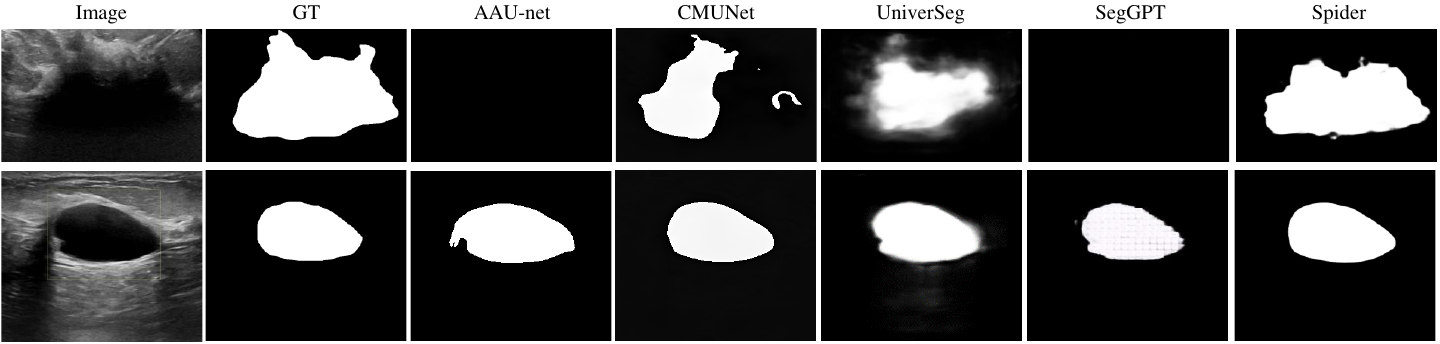}
	\caption{Qualitative comparisons with Breast Lesion Segmentation methods. AAU-net~\cite{AAU-net} and CMUNet~\cite{CMUNet} are the specialized models.}
	\label{fig:breast_comparison}
\end{figure}

\begin{figure}[!t]
	\centering
	\includegraphics[width=\linewidth]{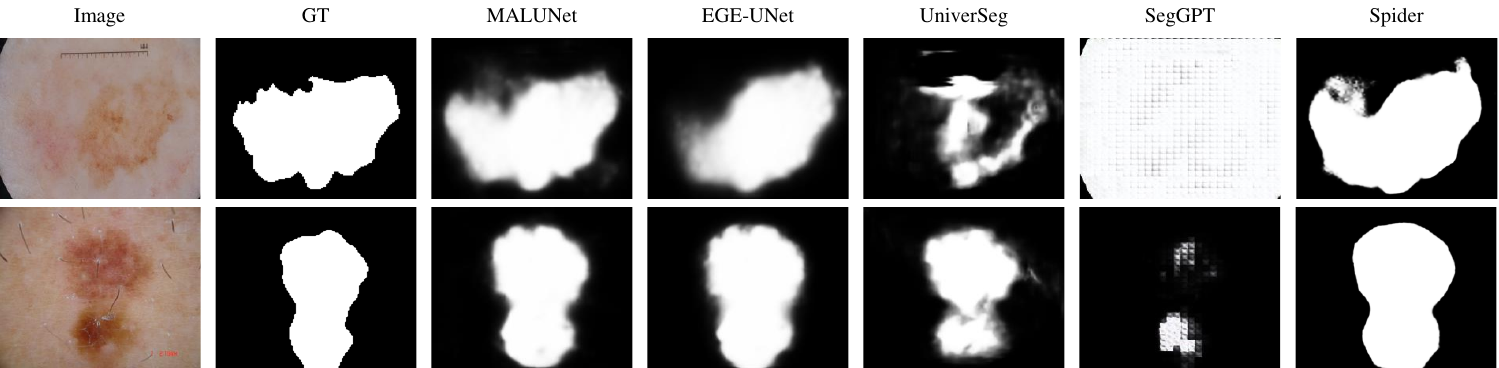}
	\caption{Qualitative comparisons with Skin Lesion Segmentation methods. MALUNet~\cite{MALUNet} and EGE-UNet~\cite{EGE-UNet} are the specialized models.}
	\label{fig:skin_comparison}
\end{figure}

In \cref{fig:sod_comparison} - \cref{fig:skin_comparison}, we show a qualitative comparison with other methods. We can see that the previous generalist model SegGPT can only distinguish the foreground and background based on the shape cues and cannot truly understand the context-dependent concept.

\subsection{Performance Analysis of Spider in Continual/Zero-shot/Incremental Zero-shot learning} 
\label{sec:IZSL}
As stated in~\cite{IZSL}, \textbf{Zero-shot learning (ZSL)} is a hot topic in transfer learning, which handles the issue that some test classes are not included in the training set.
\textbf{Continual Learning (CL)}, also known as Incremental Learning, Life-long Learning, requires the model to accumulate the knowledge of previous tasks and capture the knowledge of current tasks simultaneously. Catastrophic forgetting is the main reason why the trained model forgets the knowledge of previous tasks when a new task arrives.
\textbf{Incremental Zero-shot learning (IZSL)} is different from traditional CL and introduces the zero-shot setting. The model trained on the pervious tasks is finetuned on the new task as in CL, but tested on other unseen test sets as in ZSL. In \cref{fig:continual_learning}, based on the same model pre-trained on the first four tasks T1 - T4, five experimental settings are listed in the \cref{tab:ZSL_IZSL_CL_setting}.

\noindent\textbf{ZSL:} In S.0, directly tested on four unseen tasks T5 - T8, our performance (Shadow: 0.1732, Transparent: 0.1475, Breast: 0.4895, Skin: 0.6387) is close to or even exceeds that of existing expert models in \cref{tab:comparison}, such as EBLNet: 0.1383, AAU-net: 0.4745.

\noindent\textbf{CL:} In S.4, our method achieves the gains of -4.7\% (average performance of T1 - T4), + 63\% (T5), + 59\% (T6), + 63\% (T7), and + 44\% (T8) relative to the results in S.0. Our model has a tolerable performance degradation on old tasks and significant gains on finetuned new tasks.

\noindent\textbf{IZSL:} 
For T6, the model in S.0 has a BER of 0.1475. The model in S.1 has a BER of 0.1260 which achieves a relative improvement of 15\%. We can see that a significant performance improvement can be accomplished for T6 when implementing continuous learning once. For T8, mDice scores are 0.6381, 0.6424, and 0.7074 for the models in S.0, S.1, and S.2, respectively. The results in S.1 are almost unchanged with respect to those in S.0, and the results in S.2 show a relative improvement of 11\%. We can see that a significant performance improvement can be accomplished for T8 when implementing continuous learning twice. 

We found that the data-level correlation between old and new tasks may affect the performance of IZSL. Learning a new task once does not always immediately show improved performance on future tasks. It is important for IZSL to conduct multiple continuous learning processes to accumulate data diversity for model learning.

\begin{table*}[t]
	\centering
	\caption{Five experimental settings on Zero-shot learning (ZSL), Continual Learning (CL) and Incremental Zero-shot learning (IZSL). T1 - T8 are consistent with the expression in \cref{fig:continual_learning}.
	}
	\resizebox{0.6\linewidth}{!}{
		\setlength\tabcolsep{2pt}
		\renewcommand\arraystretch{1.2}
		 \begin{tabular}{c|c|c|c|c}
\toprule[2pt]
 Setting&Finetuning&Test Tasks for ZSL&Test Tasks for CL&Test Tasks for IZSL\\
 \hline
 S.0&-&T5 - T8&-&-\\
 S.1& S.0 + T5&-&T1 - T5&T6, T7, T8\\
 S.2& S.1 + T6&-&T1 - T6&T7, T8\\
 S.3& S.2 + T7&-&T1 - T7&T8\\
 S.4& S.3 + T8&-&T1 - T8&-\\
\bottomrule[2pt]
\end{tabular}

	}
	
	\label{tab:ZSL_IZSL_CL_setting}
\end{table*}

\subsection{Capability of In-Context Learning}
\label{sec:cap_in_context}
\begin{figure}[!t]
	\centering
	\includegraphics[width=\linewidth]{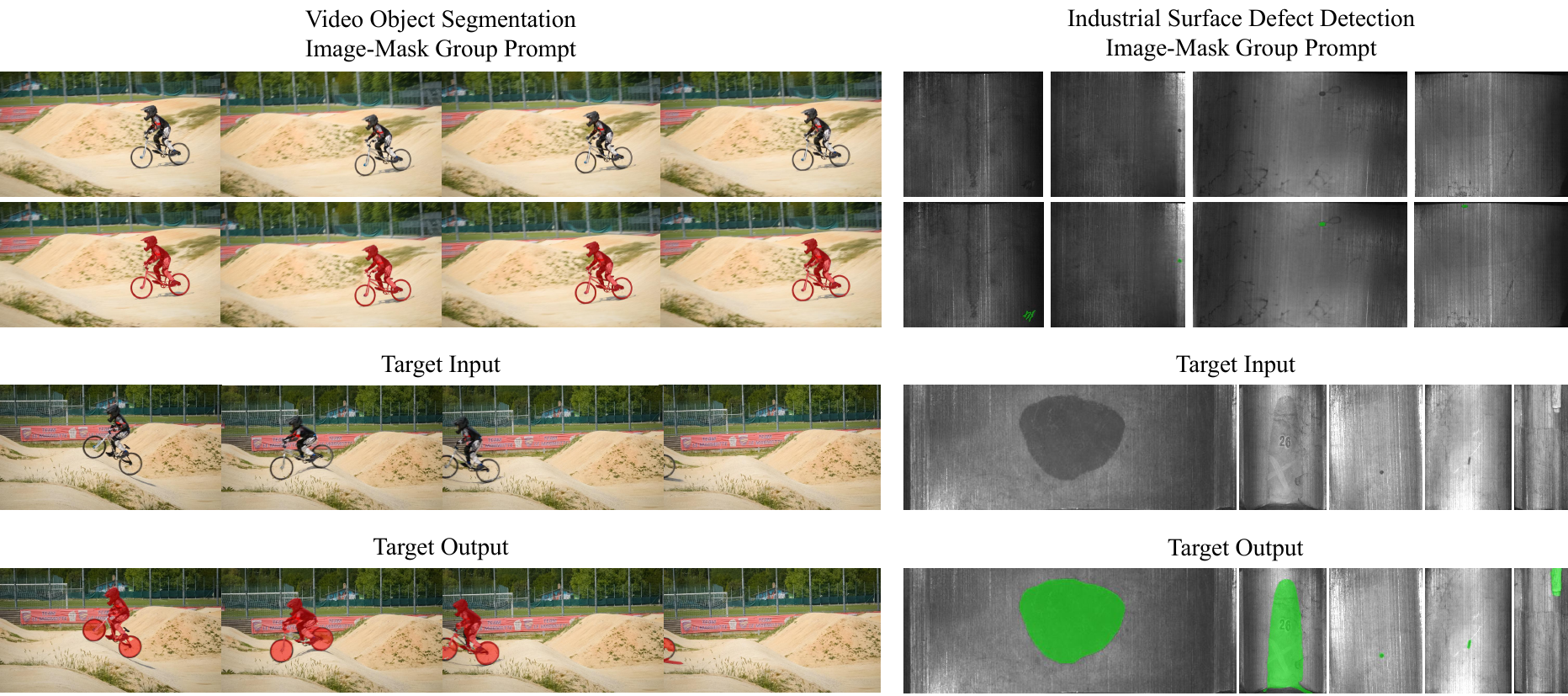}
	\caption{Visual results on the video object segmentation and industrial surface defect detection tasks.}
	\label{fig:in-context}
	\vspace{-5mm}
\end{figure}

In-context learning usually refers to completing predictions on untrained tasks and samples by providing some prompts to the model. In \cref{fig:in-context}, we separately provide our Spider with image-mask group prompts on video object segmentation and industrial surface defect detection tasks. It can be seen that Spider can capture the specified types of moving objects and defect areas. Therefore, we believe that Spider can have more powerful in-context learning capabilities with the increasing of data scale and diversity.

\end{document}